\documentclass[conference]{IEEEtran}
\IEEEoverridecommandlockouts
\usepackage{amsmath,amsfonts}
\usepackage{algorithmic}
\usepackage{algorithm}
\usepackage{array}
\usepackage{textcomp}
\usepackage{stfloats}
\usepackage{url}
\usepackage{verbatim}
\usepackage{graphicx}
\usepackage{cite}
\usepackage{xcolor}
\usepackage{subcaption}
\usepackage{float}
\usepackage{multirow}
\usepackage{makecell}
\usepackage{tikz}
\usepackage{pgfplots}
\usepackage{soul}
\usepackage{color, colortbl}
\usepackage{url}
\usepgfplotslibrary{statistics}
\DeclareUnicodeCharacter{2212}{\ensuremath{-}}
\def\BibTeX{{\rm B\kern-.05em{\sc i\kern-.025em b}\kern-.08em
    T\kern-.1667em\lower.7ex\hbox{E}\kern-.125emX}}

\newcommand{\mydiamond}{%
  \sbox0{$\lozenge$}%
  \usebox0\kern-.5\wd0\clap{\raisebox{.1ex}{\scalebox{.7}[1]{$-$}}}\kern.5\wd0%
}

\begin{document}

\title{PolyLUT: Ultra-low Latency Polynomial Inference with Hardware-Aware Structured Pruning
}

\author{\IEEEauthorblockN{Marta Andronic, Jiawen Li, George A. Constantinides}
\IEEEauthorblockA{Department of Electrical and Electronic Engineering \\
Imperial College London, UK\\
Email: \{marta.andronic18, jiawen.li21, g.constantinides\}@imperial.ac.uk}
}

\maketitle

\begin{abstract}
Standard deep neural network inference involves the computation of interleaved linear maps and nonlinear activation functions. Prior work for ultra-low latency implementations has hardcoded these operations inside FPGA lookup tables (LUTs). However, FPGA LUTs can implement a much greater variety of functions. In this paper, we propose a novel approach to training DNNs for FPGA deployment using {\em multivariate polynomials} as the basic building block. Our method takes advantage of the flexibility offered by the soft logic, hiding the polynomial evaluation inside the LUTs with minimal overhead. By using polynomial building blocks, we achieve the same accuracy using considerably fewer layers of soft logic than by using linear functions, leading to significant latency and area improvements. LUT-based implementations also face a significant challenge: the LUT size grows exponentially with the number of inputs. Prior work relies on \textit{a priori} fixed sparsity, with results heavily dependent on seed selection. To address this, we propose a structured pruning strategy using a bespoke hardware-aware group regularizer that encourages a particular sparsity pattern that leads to a small number of inputs per neuron. We demonstrate the effectiveness of PolyLUT on three tasks: network intrusion detection, jet identification at the CERN Large Hadron Collider, and MNIST.
\end{abstract}


\section{Introduction and Motivation}
Deep learning has emerged as a powerful method for deriving abstract features from raw data, achieving notable success in areas like image classification and natural language processing~\cite{goodfellow}. Deep neural networks (DNNs) are often used to analyze vast datasets, recognize patterns, and make informed decisions akin to human cognition. Ultra-low latency DNN inference on the edge has enhanced advancements in particle physics\cite{fahim}, cybersecurity\cite{murovic}, and X-ray classification\cite{cookiebox}. The migration of the models to the edge allows for rapid processing and decision-making, contributing to enhanced physics experiments and improved security systems. However, the high computational costs and stringent power and memory demands of these models pose significant challenges~\cite{esurvey}.

To satisfy the requirements of such applications, field-programmable gate arrays (FPGA) accelerators are engineered through hardware/software co-design\cite{finn, fahim}. FPGAs are an ideal platform for designing ultra-low latency deep learning accelerators due to their high customizability, parallel processing capabilities, reconfigurability, energy efficiency, deterministic performance, and high data throughput. To reduce the high computational cost of deploying neural networks on the edge, common compression techniques such as pruning and quantization are also employed. 

Binary neural networks (BNNs) represent an extreme form of quantization, reducing both weights and activations to only two values: +1 and -1. Despite their simplicity, BNNs can achieve good accuracy levels while significantly reducing the computational cost by replacing resource-intensive dot products with efficient XNOR and bitcount operations. This reduction in computational complexity, coupled with drastic memory savings, makes BNNs particularly well-suited for deployment in resource-constrained environments such as edge devices~\cite{edge}.

However, BNNs still have limitations when implemented on FPGAs because they do not fully utilize the flexibility of the FPGA fabric. Specifically, some XNOR and bitcount operations get implemented with $K$-LUTs, which are capable of implementing arbitrary Boolean functions with limited support. Previous works that leverage LUTs beyond simple XNOR operations fall into two categories: Differentiable LUTs (such as LUTNet~\cite{lutnet1}) and LUT-based traditional networks (like LogicNets~\cite{logicnets} and NullaNet~\cite{nullanet}).

LUTNet~\cite{lutnet1} exploits the flexibility of the $K$-LUTs and replaces BNNs' XNOR operations with learned $K$-input Boolean operations~\cite{lutnet1}. By using LUTs as inference operators, LUTNet achieves high logic density, enabling extensive pruning. However, the number of trainable parameters grows exponentially with the size of the LUT inputs.

Similar to prior BNN works, LUTNet is bottlenecked by the exposed datapaths inside a neuron, for example the adder trees. To combat this, LogicNets and NullaNet encapsulate the full computation of a traditional neuron within a $N\text{-input}$ Logical-LUT which gets synthesized by the back-end tools to $K\text{-input}$ Physical-LUTs. In the case where $N \leq K$ the Logical-LUT gets implemented to a single Physical-LUT.

Given that the neural network is encapsulated in a netlist of LUTs, we propose leveraging this structure to compute more sophisticated functions, thereby maximizing the computational complexity within these LUTs. Traditional neurons typically compute linear functions followed by rectified linear units (ReLUs), resulting in the overall neural network computing a continuous piecewise linear function. Instead of using these simple functions, we propose training the network to compute continuous piecewise polynomial functions. This approach enhances the network's learning capacity while keeping the truth table sizes fixed.

By performing polynomial transformations within each neuron, we augment the feature space with all monomials up to a tunable degree
$D$. This provides a flexible trade-off between model generalization capability and the number of training parameters. Consequently, this method enables the creation of significantly shallower and lower-latency networks, which are highly efficient for real-time applications and resource-constrained environments.

The L-LUT size scales exponentially limiting network architecture with few inputs, each of low precision. Since the number of inputs is restricted to a small value, the input selection process has proven to be a significant factor in the performance and robustness of neural networks. Various methods have been proposed to limit the fan-in of each neuron, using techniques such as lossy function approximations\cite{nullanet} and random sparsity maps\cite{logicnets}, however, we show that they introduce high stochasticity in the final accuracy. High stochasticity diminishes the predictability of the neural network, leads to non-reproducible results, and can affect the performance of algorithms such as automatic hyperparameter tuning or neural architecture search.

We propose integrating a structural sparsification schedule into PolyLUT. This schedule entails an initial stage of dense training using a novel hardware-aware regularizer, which was specifically designed for this type of neural network. The dense training is followed by structured pruning after which the sparse network undergoes retraining. It is important to note that we restrict the initial stage of pruning to just a few epochs (for example, $25$), ensuring minimal impact on training time.

Our aim is to support applications demanding ultra-low latency real-time processing and extremely lightweight on-chip implementations. Therefore, our target applications are classifying particle collision events~\cite{duarte}, detecting malicious network traffic~\cite{murovic}, and performing image classification for real-time systems. Consequently, similar to other research in this area, we concentrate on fully-unrolled networks that can be implemented on a single device.

\begin{figure*}[!b]
     \centering
     \hspace{-9mm}
     \begin{subfigure}[b]{0.3\textwidth}
         \captionsetup{width=.9\textwidth}
         \includegraphics[width=\textwidth]{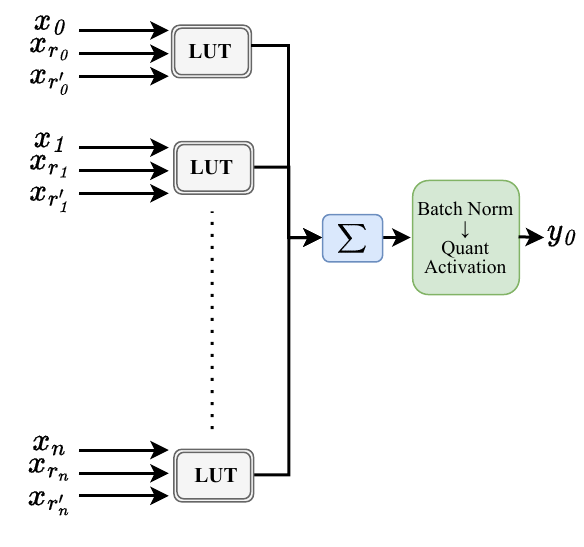}
         \caption{LUTNet - BNN's XNOR operations replaced with $K$-LUTs, here $3$-LUTs. $x$ would be the input vector to the XNORs. $x_r$ and $x_{r'}$ are random selections (with replacement) from $x$.}
         \label{fig:lutnet}
     \end{subfigure}
     \hspace{4mm}
     \begin{subfigure}[b]{0.28\textwidth}
         \captionsetup{width=.8\textwidth}
         \includegraphics[width=\textwidth]{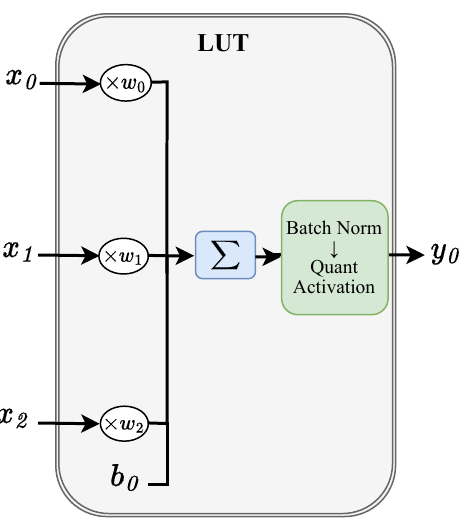}
         \caption{LogicNets - L-LUTs are used to absorb all the operations between the quantized inputs and quantized output.\\}
         \label{fig:logicnets}
     \end{subfigure}
     \hspace{3.5mm}
     \begin{subfigure}[b]{0.36\textwidth}
         \captionsetup{width=.75\textwidth}
         \includegraphics[width=\textwidth]{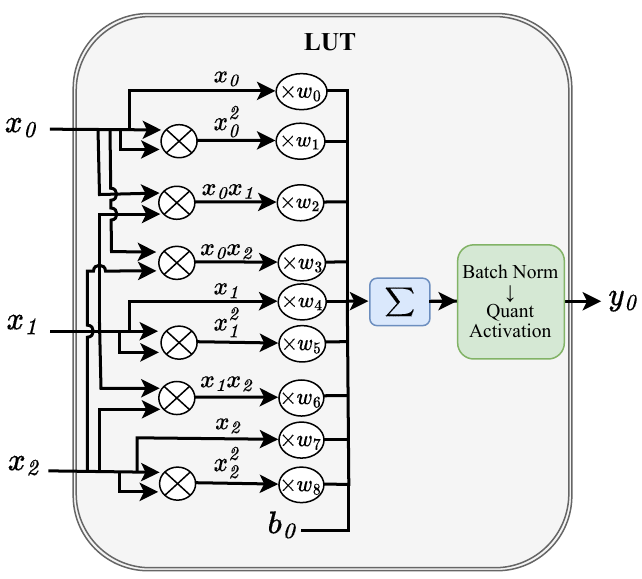}
         \caption{PolyLUT - L-LUTs are used similar to LogicNets, however, inside the neuron a multivariate polynomial of degree $D$ (here $D=2$) gets computed.}
         \label{fig:polylut}
     \end{subfigure}
     \hspace{-8mm}
        \caption{Structural use of LUTs for a single output channel in prior works (a,b) and our approach (c).}
        \label{fig:three graphs}

\end{figure*}

The main novel contributions of this work are as follows:
\begin{itemize}
    \item{We introduce PolyLUT, an open-source framework and the first quantized neural network training methodology that maps neurons capable of computing multivariate polynomials to logical-LUTs to harness the flexibility of FPGA soft logic\footnote{\url{https://github.com/MartaAndronic/PolyLUT}}}.
    \item{We expose the degree of the monomials as a training hyperparameter that can be adjusted to provide full control of the flexibility of the functions. We demonstrate empirically that higher degrees increase the ability of the model to fit the training data which allows significant network depth reductions.}
    \item{We perform hardware-aware structured pruning and propose a hardware-aware group regularizer to promote structured sparsity. Empirical evidence demonstrates that, after retraining, the model's accuracy increases compared to prior works, and the standard deviation of accuracy across different runs is reduced by a factor of $6.7\times$.}
    \item{We show that PolyLUT maintains accuracy on shallower models, resulting in Pareto optimal designs both in terms of both latency area with respect to test error rate. On jet substructure tagging, a 2-layer piecewise polynomial network achieves the same accuracy as a 4-layer piecewise linear network, thus reaching $2.69\times$ area compression and $2.29\times$ latency reduction.
    }
\end{itemize}

This work forms an extension of an earlier conference paper~\cite{poly}; the key differences are the introduction of a novel hardware-aware regularization term and the replacement of the random-pruning strategy with a structured-pruning pipeline (Section~\ref{section:reguralizer}). These improvements not only enhance test accuracy, Section~\ref{section:ablation}, but also significantly reduce the stochastic variability of the test results observed in the prior implementations, Section~\ref{section:stochastic}.

\section{Background}
Unlike cloud computing, deploying DNN models on edge devices poses substantial challenges because of their constrained computational resources~\cite{edge}. To overcome these challenges, research efforts have focused on compression techniques, algorithm-hardware co-design, and the use of hardware accelerators. Table~\ref{table:prior} provides a summary of techniques utilized in previous low-latency FPGA accelerators.

\subsection{XOR-based Neural Networks}
FINN~\cite{finn}, an open-source framework initially developed for building high-performance BNN accelerators on FPGAs, reported the fastest classification rates on MNIST at the time of the publication. FINN has proposed several hardware-specific optimizations such as replacing additions with popcount operators, batch normalizations and activation functions with thresholding, and max-pooling with Boolean ORs. In parallel, \texttt{hls4ml}~\cite{hls4ml} emerged as an open-source framework focused on mapping neural networks to FPGAs for ultra-low latency applications. Ngadiuba \textit{et al.}\cite{hls4ml} leveraged \texttt{hls4ml} to implement fully unrolled binary and ternary neural networks, aiming to reduce computation time and resource usage. Additionally, Murovic \textit{et al.}~\cite{murovic} presented a real-time BNN implementation achieving high performance through full parallelization. However, these BNN works fully focus on binarization and do not leverage any pruning techniques, implementing the full network on the device.

\subsection{DSP-based Neural Networks}
Duarte \textit{et al.}\cite{duarte} employed \texttt{hls4ml} with higher network precision to produce accurate and fast fully-unrolled and rolled designs, however this has resulted in significant DSP block usage. Consequently, magnitude-based iterative pruning was utilised with $l_1$ regularization to achieve a $70\%$ model compression. Fahim \textit{et al.} extended the prior \texttt{hls4ml} works by integrating techniques from previous studies, such as boosted decision trees~\cite{summers} and quantization-aware training~\cite{coelho}, and introduced quantization-aware iterative pruning to enhance performance and power efficiency. Quantization-aware pruning involves pruning weights during the quantization-aware training, as opposed to the prior methods of pruning full-precision weights followed by quantization.

\begin{table*}[tbp]
\caption{Key features of previous FPGA low-latency accelerators compared to our work.}
\begin{center}
\renewcommand{\arraystretch}{1.2}
\setlength{\tabcolsep}{14pt}
\begin{tabular}{clll}
\Xhline{2\arrayrulewidth}
\multirow{2}{*}{\textbf{Tool}}&\multirow{2}{*}{\textbf{Precision}}&\multirow{2}{*}{\textbf{Pruning type}}&\multirow{2}{*}{\textbf{Implementation type}}\\&&&\\
\Xhline{2\arrayrulewidth}
\multirow{2}{*}{FINN~\cite{finn}}&\multirow{2}{*}{Binary}&\multirow{2}{*}{None}&\multirow{2}{*}{XNOR-based}\\&&&\\
\hline
\multirow{2}{*}{Ngadiuba \textit{et al.}~\cite{hls4ml}}&\multirow{2}{*}{Binary / Ternary}&\multirow{2}{*}{None}&\multirow{2}{*}{XNOR-based}\\&&&\\
\hline
\multirow{2}{*}{Murovic \textit{et al.}~\cite{murovic}}&\multirow{2}{*}{Binary}&\multirow{2}{*}{None}&\multirow{2}{*}{XNOR-based}\\&&&\\
\hline
\multirow{2}{*}{Duarte \textit{et al.}~\cite{duarte}}&\multirow{2}{*}{Fixed point}&\multirow{2}{*}{Iterative pruning}&\multirow{2}{*}{DSP-based}\\&&&\\
\hline
\multirow{2}{*}{Fahim \textit{et al.}~\cite{fahim}}&\multirow{2}{*}{Fixed point}&Quantization-aware&\multirow{2}{*}{DSP-based}\\&&iterative pruning&\\

\hline
\multirow{2}{*}{LUTNet \cite{lutnet1}}&\multirow{2}{*}{Residual binarization}&\multirow{2}{*}{Iterative pruning}&\multirow{2}{*}{LUT-based}\\&&&\\
\hline
\multirow{2}{*}{LogicNets \cite{logicnets}}&\multirow{2}{*}{Custom low-bit}&\multirow{2}{*}{\textit{A priori} fixed sparsity}&\multirow{2}{*}{LUT-based}\\&&&\\
\hline
\multirow{2}{*}{PolyLUT (Ours)}&\multirow{2}{*}{Custom low-bit}&Quantization-aware&\multirow{2}{*}{LUT-based}\\&&iterative pruning&\\
\end{tabular}
\renewcommand{\arraystretch}{1.2}
\label{table:prior}
\end{center}
\end{table*}
\subsection{BNN-specific micro-architectures using FPGA LUTs}
\subsubsection{Differentiable LUTs} LUTNet substitutes the dot product of the feature vector and weight vector with a sum of learned Boolean functions, each implemented using a single $K$-LUT. The architecture of LUTNet is depicted in Figure~\ref{fig:lutnet}, where $K=3$ and $x_r$ and $x_{r'}$ are vectors created by randomly selecting (with replacement) from the input vector $x$. This approach allows each input to contribute multiple times to the weighted summation, and redundant operations can be minimized through network pruning while preserving accuracy~\cite{lutnet1}.

In contrast to PolyLUT, LUTNet features exposed datapaths within neurons, which give rise to bottlenecks, it shows exponential growth in training parameters with the size of LUT inputs, and it supports only residual binarization.

\subsubsection{LUT-based traditional networks} NullaNet~\cite{nullanet} and LogicNets~\cite{logicnets} treat layers as multi-input multi-output Boolean functions. In NullaNet, on top of optimizing the Boolean functions via Boolean logic minimization, output values are specified only for a strict subset of input combinations in order to manage the exponential increase in truth table size. Therefore, the logic synthesis tools further optimize the circuit by treating the remaining combinations as \textit{don't-care} conditions. In contrast, LogicNets~\cite{logicnets} employs high sparsity to mitigate the issue of large fan-in, countering the drawback of NullaNet’s lossy truth table sampling approach. Figure~\ref{fig:logicnets} illustrates the structure of a LogicNets neuron. In a dense implementation, the neuron would have $N$ inputs, however, due to sparsity, the input vector $x$ has a size $F \ll N$, with $F=3$ in the figure. Post-training, the function incorporated into the L-LUT is expressed as shown in (\ref{logicnets_lut}), where $\phi$ represents the quantized activation function, $w$ is the real weight vector, and $b$ is the real bias term.
\begin{equation}
\label{logicnets_lut}
y_0=\phi\left[\sum_{i=0}^{F-1}w_ix_i + b\right]
\end{equation}

\subsection{Pruning}
Network pruning is utilized to achieve high sparsity while maintaining accuracy, demonstrating a regularization effect that enhances neural network generalization. Remarkably, this technique has been shown to maintain accuracy despite a $9\times$ reduction in the number of parameters, as evidenced by results on the ImageNet dataset~\cite{songhan}.

The process of network pruning typically follows a three-stage pipeline. The first stage involves training a dense network, the second stage entails pruning the smallest weights in magnitude, and the final stage focuses on retraining the sparse network. In some cases, this can be an iterative process involving multiple cycles of pruning and retraining to achieve optimal sparsity and performance.

Regularization plays a crucial role in the initial training stage. The most widely used weight regularization techniques in neural networks are the $l_2$ and $l_1$ regularizers. These methods are essential for preventing overfitting by penalizing large weights, thereby encouraging simpler and more generalizable models. The $l_2$ regularizer achieves this by adding a penalty proportional to the sum of the squares of the weights, promoting uniform weight shrinkage. In contrast, the $l_1$ regularizer adds a penalty proportional to the absolute value of the weights, leading to sparser models by driving some weights to zero.

However, these traditional network pruning techniques result in unstructured sparsity, which is inefficient for hardware implementation, resulting in limited parallelism and irregular memory access patterns. To address this issue, structured pruning can be employed, although traditional regularizers may not be as effective. A previously proposed regularizer that encourages structure uses Group Lasso~\cite{glasso}, as shown in Equation~\ref{group_lasso}, where $\lambda$ is the weight decay parameter and $W_i$ are the weights corresponding to one group. This method selects or discards entire groups of weights together, which can be useful for eliminating entire neurons when a group represents a neuron. However, in LUT-based neural networks, the goal is to retain all neurons while encouraging high sparsity at the input of each neuron. Therefore, to achieve this goal we propose a new regularizer, which will be discussed in Section~\ref{section:reguralizer}.
\begin{equation}
\label{group_lasso}
    \Omega(W) = \lambda \sum_{i=1}^{G} \| W_i \|_{2}^2
\end{equation}

\section{Methodology}
\label{section:methodology}
Our key contribution is the replacement of each feature vector $x$ as seen in Equation~\ref{logicnets_lut} with an expanded vector $x$ which contains all the monomials up to a parametric degree, $D$. Figure~\ref{fig:polylut} illustrates PolyLUT's expansion for $D=2$. When $D=1$, PolyLUT behaves linearly, equivalent to a network like LogicNets. It is thus a strict generalization of LogicNets.

\subsection{Theoretical approach}

Given a sufficient depth and width, a deep neural network is capable of modelling any continuous function within a specific error margin~\cite{hornik}. There is theoretical support that multiplicative interactions enhance the neural network with contextual information~\cite{multi}. Moreover, it is shown that the hypothesis space of a conventional multi-layer perceptron with rectified linear unit (ReLU) activation functions is strictly contained within the hypothesis space of an equivalent network where each linear layer is replaced by a layer with multiplicative interactions~\cite{multi}. Hence, these multiplicative interactions expand the hypothesis space, increasing representational power, interaction modeling, and robustness.

In a direct hardware implementation, incorporating multiplicative terms in the DNN would increase the resource utilisation significantly, {\em however, in a LUT-based model, this additional complexity is entirely absorbed inside the logical-LUT, and no additional multiplication hardware is required}. This is the key to our approach, leading to a significant boost in model performance with minimal overhead. We leverage the increase in function complexity within each layer to reduce the number of layers needed to achieve a given accuracy.

Our modified layers are fully-trainable and they can be described in the following way. Consider a vector $x$ of inputs to a standard neural network layer. We may first form all multiplicative combinations of these inputs up to a user-defined degree $D$. For example, if a given input vector is two-dimensional and $D=3$, then model construction arises from the following transformation: $[x_0,x_1]\mapsto[1,x_0,x_1,x_0^2,x_0x_1,x_1^2,x_0^3,x_0^2x_1,x_0x_1^2,x_1^3]$. The size of the feature vectors will see an increase from $F$ to $\frac{(F+D)!}{F!D!}$, where $D$ can be seen as a hyperparameter allowing a smooth variation from a number of terms linear in $F$ when $D=1$, equivalent to LogicNets, to a number of terms exponential in $F$ when $D=F$, akin to LUTNet, as seen in Table~\ref{table:params}. After this transformation, we apply a standard linear + activation transformation to the resulting expanded feature vector. This can therefore be seen as replacing (\ref{logicnets_lut}) with (\ref{polylut_lut}), where $M$ is equal to the number of monomials $m(x)$ of degree at most $D$ in $F$ variables.
\begin{equation}
\label{polylut_lut}
y_0=\phi\left[\sum_{i=0}^{M}w_im_i(x)\right]\text{, where M} = \left(F+D \atop D \right)
\end{equation}

\begin{table}[t]
\caption{Number of parameters for a $\beta F$-input LUT, where $\beta$ is the input bitwidth and $F$ is the neuron fan-in. PolyLUT's parameter count falls between that of LogicNets and LUTNet. Unlike LUTNet, the parameter count for LogicNets and PolyLUT is independent of the input bitwidth.}
\begin{center}
\renewcommand{\arraystretch}{1.5} 
\begin{tabular}{ccc}
\Xhline{2\arrayrulewidth}
&\multirow{2}{*}{\textbf{Number of parameters}}&\multirow{2}{*}{\textbf{Scaling type}}\\&&\\
\Xhline{2\arrayrulewidth}
\multirow{2}{*}{\textbf{LUTNet}}&\multirow{2}{*}{$2^{\beta F}$}&Exponential in F\\&&for fixed $\beta$\\
\hline
\multirow{2}{*}{\textbf{PolyLUT}}&\multirow{2}{*}{$\left(F+D \atop D \right)$}&Polynomial in F\\&&for fixed D\\
\hline
\multirow{2}{*}{\textbf{LogicNets}}&\multirow{2}{*}{$F+1$}&\multirow{2}{*}{Linear in F}\\&&\\
\hline

\end{tabular}
\label{table:params}
\end{center}
\end{table}

ReLU is a widely used activation function in deep learning. Functionally, each ReLU partitions the input space into two linear sub-regions, introducing non-linearity by outputting the input directly if it is positive, and zero otherwise. As the network depth increases, the composition of multiple ReLU activations generates more piecewise linear surfaces, enhancing the model's capability to represent sophisticated data distributions.

In contrast, the polynomial setting transforms the input space into a surface described by a multivariate polynomial function. This transformation provides significantly greater flexibility and expressive power than linear transformations alone. Networks incorporating polynomial transformations can accurately capture complex relationships in the data, utilizing fewer neurons or layers. This is because polynomial functions can represent a wider variety of shapes and interactions within the data, enabling the network to achieve high accuracy with reduced depth and computational resources.

\begin{figure}[!t]
\centerline{\includegraphics[width=90mm]{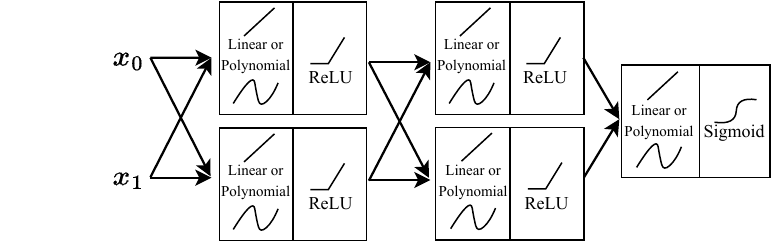}}
\caption{Illustration of a toy three-layer neural network.}
\label{fig:net}
\end{figure}

\begin{figure}[!t]
     \centering
     \begin{subfigure}[b]{0.49\textwidth}
         \includegraphics[width=\textwidth]{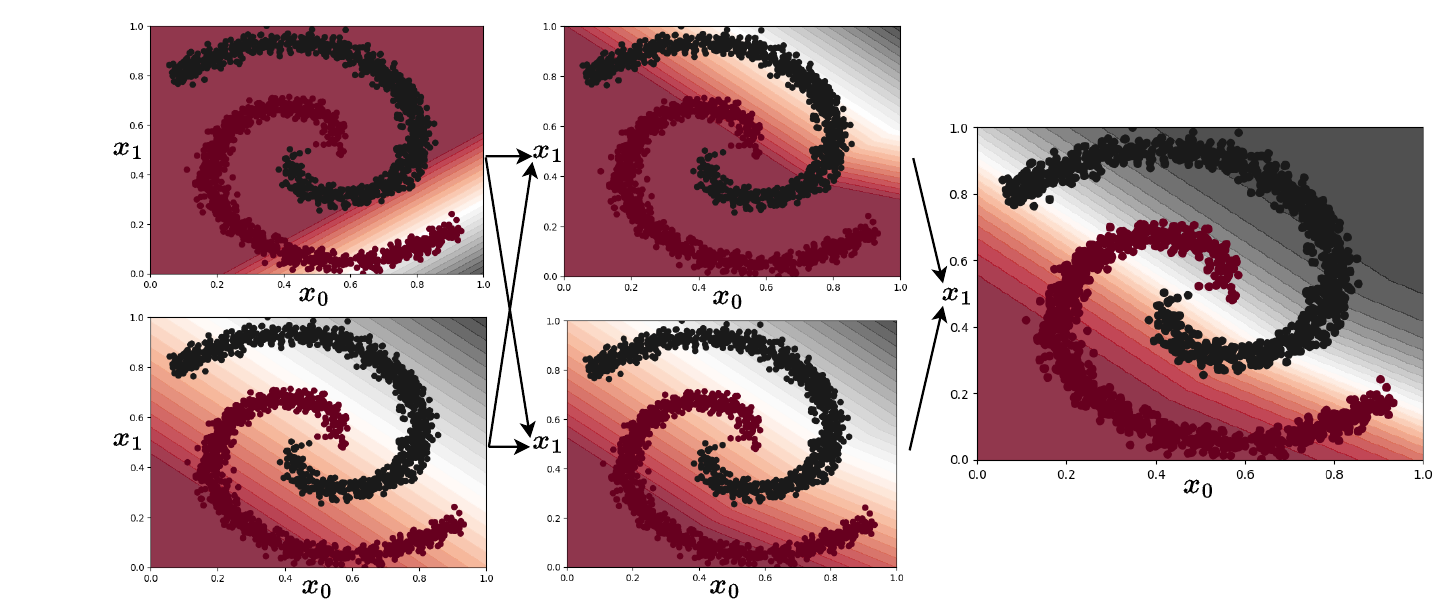}
         \caption{The formation of a continuous piecewise linear decision boundary.}
         \label{fig:linear}
     \end{subfigure}
     \hspace{2mm}
     \begin{subfigure}[b]{0.49\textwidth}
         \includegraphics[width=\textwidth]{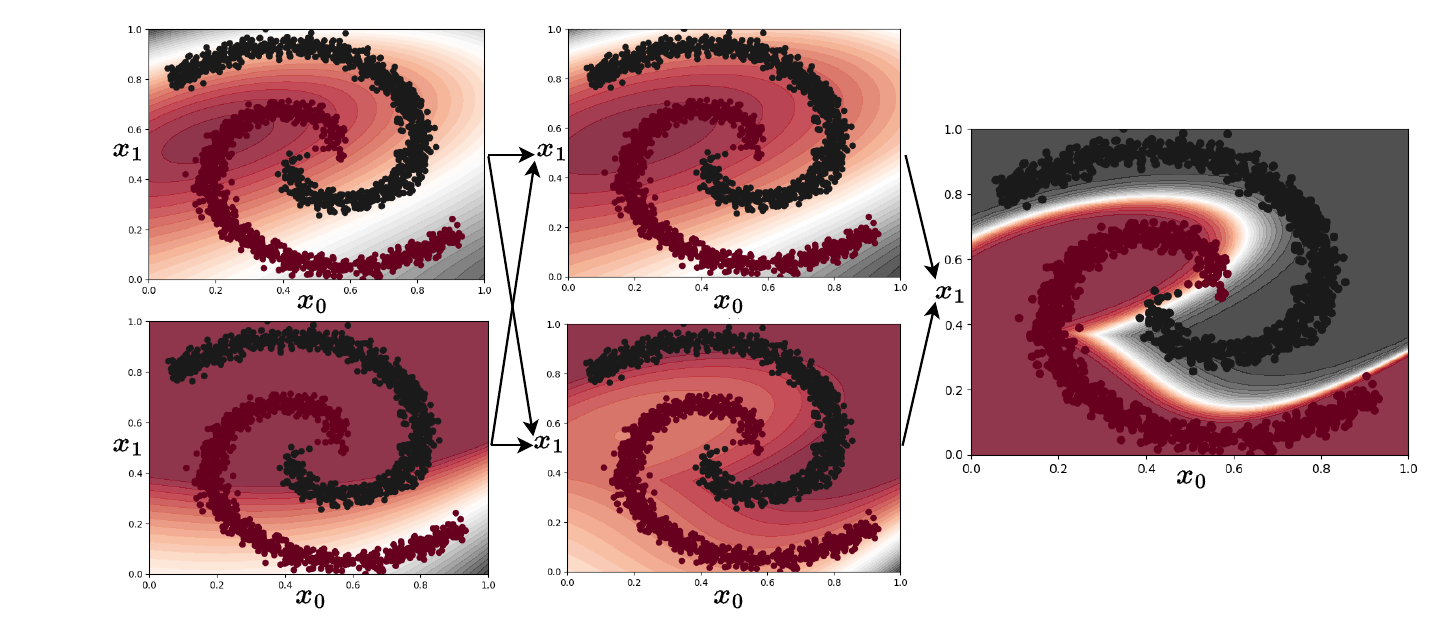}
         \caption{The formation of a continuous piecewise polynomial decision boundary.}
         \label{fig:poly}
     \end{subfigure}
        \caption{Input transformations visualized as contour graphs at the output of each neuron. The black and red dotted spirals are the datapoints used for training.}
        \label{fig:two graphs}
\end{figure}
To illustrate the advantage of using piecewise polynomial functions in neural networks, we consider a simple two-dimensional example. The toy network in Figure~\ref{fig:net} is trained on a fictional dataset comprising two intertwined spirals, each representing a different class (black and red).

Initially, we train the network using linear functions. In this setup, each neuron applies a linear transformation to its input, followed by a ReLU activation function. The ReLU function introduces non-linearity by partitioning the input space into two linear regions. As the network processes the data, the first layer of neurons attempts to separate the spirals using a single linear transformation. Subsequent layers add more linear transformations, creating multiple linear regions. This results in a piecewise linear surface that attempts to classify the data, as visualized in Figure~\ref{fig:linear}. However, due to the complexity of the intertwined spirals, the network's classification accuracy is limited by the insufficient flexibility of linear functions. To improve accuracy, the network's width or depth would need to be increased, which also increases computational complexity.

Next, we enhance the network by using polynomial functions of degree three. Instead of linear transformations, each neuron now applies a polynomial transformation to its input. This change significantly increases the network's ability to capture complex data patterns. The resulting piecewise polynomial surface can adopt more intricate shapes, allowing the network to better distinguish between the intertwined spirals. This improved data fitting capability leads to higher classification accuracy with fewer neurons and layers, reducing computational complexity and latency, as shown in Figure~\ref{fig:poly}.

However, using high-degree polynomials introduces more degrees of freedom, which can be harder to train or increase the risk of overfitting. To address this, two regularization techniques can be used: decreasing the polynomial degree or reducing the number of layers. Decreasing the degree maintains generalization with fewer parameters, while reducing the number of layers limits the number of polynomial regions and leads to a reduction in the clock cycles per classification, saving computational resources and improving efficiency.

\subsection{Hardware approach}
A LUT can evaluate any function with quantized inputs and outputs. For a fixed number of LUT inputs, the evaluation time complexity remains constant, regardless of the function's complexity stored in the LUT. This characteristic makes LUTs highly effective for implementing computationally intensive functions, as they can significantly streamline complex computations.

The $K$-LUT is a fundamental component of FPGA soft logic, capable of implementing any Boolean function with $K$ inputs and one output. In this work, we refer to this as a P-LUT. Utilizing truth tables to represent artificial neuron computations is highly efficient, as it consolidates multiple operations—such as multiplication, summation, activation, quantization, and normalization—into a single operation that can be executed within one clock cycle or less.

However, restricting neuron inputs to $K$ would result in a significant accuracy penalty. To address this, we work with Logical-LUTs (L-LUTs), which typically have more inputs than P-LUTs and are implemented by synthesis tools as circuits composed of multiple P-LUTs, in common with LogicNets. In the worst case, the number of P-LUTs required to implement a single L-LUT still scales exponentially with the number of L-LUT inputs for $L > K$. Therefore, sparsity must be enforced in the connections between L-LUTs, limiting the neuron input fan-in to some user parameter, $F$. Additionally, we apply heavy quantization, restricting the input bit-width to $\beta < 4$ bits. Consequently, the number of P-LUTs required to implement an L-LUT becomes $\mathcal{O}(2^{\beta \cdot F})$. The complexity of the truth tables is therefore established through a trade-off between data representation precision and network sparsity.

\subsection{Hardware-Aware Regularization}
\label{section:reguralizer}
\begin{figure}[t]
\centerline{\includegraphics[width=100mm]{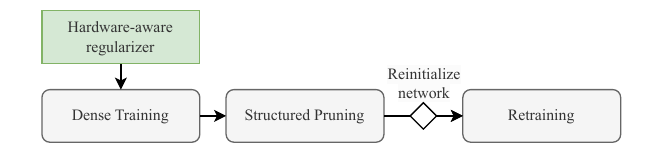}}
\caption{Three-stage network training pipeline.}
\label{fig:pipeline}
\end{figure}

Instead of achieving extreme sparsity by applying a random \textit{a priori} sparsity mask as implemented in LogicNets~\cite{logicnets}, and PolyLUT~\cite{poly}, we propose employing a pruning strategy utilizing a three-stage network training pipeline as shown in Figure~\ref{fig:pipeline}.
The first stage involves training a fully-connected network ($25$ epochs in our experiments). To encourage sparsity across the inputs of each neuron, we introduce a novel hardware-aware penalty term as seen in Equation~\ref{group_lasso}. This term exponentially penalizes the $l_1-\text{norm}$ of the weights within one group, where a group represents the collection of a dense neuron's inputs. While $\lambda_1$ is the traditional weight decay parameter, we have added an extra hyperparameter for the base of the $l_1$ norm, $\lambda_2$.
\begin{equation}
\label{group_lasso}
    \Omega(W) = \lambda_1 \sum_{i=1}^{G} \lambda_2^{||W_i||_1}
\end{equation}

\begin{figure}[t]
\centerline{\includegraphics[width=100mm]{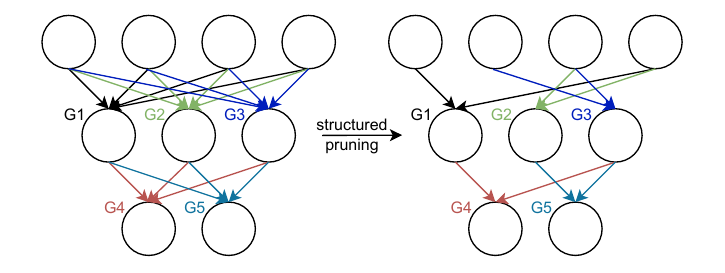}}
\caption{Structured network pruning.}
\label{fig:pruning}
\end{figure}

The second stage involves the structured pruning, ensuring that only the top \textit{k} connections from each group remain, as shown in Figure~\ref{fig:pruning}. The third stage is reinitializing the network while keeping the learned sparsity mask and retraining from scratch. This has been found to be superior compared to fine-tuning since the pruned architecture is significantly different from the original due to the extreme sparsity.

\subsection{Sources of stochasticity}

The training process for neural networks incorporates techniques that introduce stochasticity. Common sources of this stochasticity include the random initialization of weights and biases, as well as various optimization algorithms. In LUT-based neural networks, such as LogicNets\cite{logicnets} and PolyLUT~\cite{poly}, an additional source of stochasticity arises from \textit{a priori} random pruning. It is important to study these sources of stochasticity, as they can significantly affect the stability and reproducibility of neural network performance. We perform a study on this phenomenon in Section~\ref{section:stochastic}.

\subsection{Generating a neural network as a netlist of LUTs.}
\begin{figure}[t]
\centerline{\includegraphics[width=65mm]{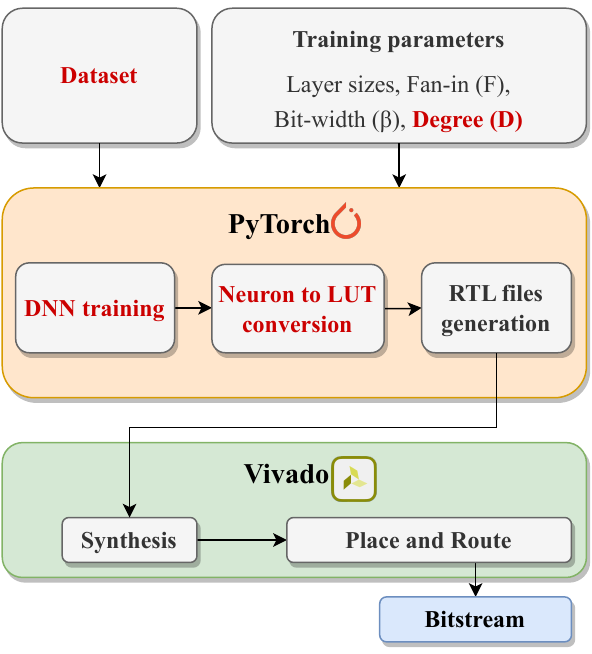}}
\caption{High-level overview of PolyLUT's toolflow, based on the open-source LogicNets toolflow. Modified elements are highlighted in red.}
\label{fig:flow}
\end{figure}

PolyLUT is a DNN co-design methodology that extends the LogicNets toolflow~\cite{logicnets}. This toolflow facilitates the quantized-aware DNN training, the model conversion into truth tables, and the generation of hardware netlists. We have modified the original LogicNets implementation to accommodate PolyLUT's features. Figure~\ref{fig:flow} provides an overview of the toolflow, with modifications highlighted in red.

\subsubsection{DNN Training}
The full training process is realized inside a PyTorch framework. Prior to training, users need to provide specific parameters, such as the dataset, model layer sizes, fan-in, bit-width, degree, and other machine learning-specific parameters like learning rate and weight decay.

A hyperparameter, \texttt{degree}, was introduced in the framework to allow users to adjust the complexity of the polynomial functions.  Setting this parameter to one enables linear learning. Our models are trained using Decoupled Weight Decay Regularization~\cite{reg} and Stochastic Gradient Descent with Warm Restarts~\cite{sgd}. Each layer's inputs and outputs are batch normalized and quantized using Brevitas~\cite{brevitas} quantized activation functions, which utilize learned scaling factors.

\subsubsection{Neuron to L-LUT conversion}
Once the model is trained, the next step is to evaluate the neuron functions to generate the truth tables. This involves enumerating all possible input combinations based on the specified bit-width and fan-in, then evaluating the neuron function on these combinations to obtain the corresponding outputs. Each neuron requires the evaluation of $2^{\beta \cdot F}$ input combinations.

\subsubsection{RTL files generation}
Each L-LUT is then converted to Verilog RTL by writing out the truth tables as read-only memories (ROMs), which are synthesized into P-LUTs by the synthesis tool. Additionally, each ROM is equipped with output registers, and the input connections are defined using activation masks.

\section{Experimental Results}
\subsection{Datasets}
To evaluate our method, we have analyzed three datasets commonly used for evaluating ultra-low latency and size-critical architectures. The first is the UNSW-NB15 dataset~\cite{unsw}, the second is the MNIST dataset~\cite{mnist}, and the final one is the jet tagging dataset as presented in~\cite{duarte}. The jet tagging dataset is also used in the case study in Section~\ref{section:case}, which focuses on the performance of learning piecewise polynomial functions. A study on the sources of stochasticity and the extent to which this is reduced with our methodology can be found in Section~\ref{section:stochastic}. Following this, we show the performance advantage of the polynomials, when combined with our pruning pipeline, on the jet substructure dataset in Section~\ref{section:ablation1}. To provide diversity, the MNIST dataset is employed in the ablation study in Section~\ref{section:ablation}, where we analyze the performance of the hardware-aware structured pruning strategy compared to the baseline and in combination with polynomials. Finally, we evaluate PolyLUT against prior work in Section~\ref{section:evaluation}.

\subsubsection{Network intrusion detection}
In cybersecurity, network intrusion detection systems (NIDS) are essential for monitoring network traffic and identifying malicious packets. With fiber-optic internet speeds reaching up to $940$ Mbps, NIDS must efficiently handle high data throughput to be effective. On-chip FPGA designs are particularly suited for this purpose, addressing both performance and privacy concerns by keeping sensitive data processing on the device.

To demonstrate our approach's effectiveness, we utilized the UNSW-NB15 dataset, which offers a comprehensive benchmark for evaluating NIDS. The dataset, partitioned randomly as described in~\cite{murovic}, includes $49$ features per network packet and labels them as either safe ($0$) or malicious ($1$).

\subsubsection{Handwritten digit recognition}
Recent advancements in autonomous driving, augmented reality, and edge technologies have increased the demand for real-time image classification. The capability to perform low-latency inference unlocks numerous applications in these rapidly evolving fields. To evaluate our work, we selected the MNIST dataset, which matches our networks' capabilities. The MNIST dataset comprises handwritten digits shown as $28\times28$ pixel images. The images are flattened into input sizes of $784$, with $10$ output classes corresponding to each digit.

\subsubsection{Jet substructure classification}
Real-time inference and efficient resource utilization have been vital for advancing physics research at the CERN Large Hadron Collider (LHC). Due to the LHC experiments' high collision frequencies, the initial data processing phase requires exceptional throughput to manage the large volumes of sensor data effectively. Previous studies~\cite{duarte,fahim,hls4ml,summers,coelho} have utilized neural networks to deploy inference models on FPGAs for jet substructure analysis. This task involves analyzing $16$ substructure properties to classify $5$ types of jets. Our work also addresses this task, demonstrating that PolyLUT can be a viable solution for high-energy physics applications.

\subsection{Training and implementation setup}
For our experiments, we trained the models on Jet Substructure and UNSW-NB15 for $1000$ epochs with a batch size of 1024. For MNIST models were trained for 500 epochs and the batch size used was $256$.

For the hardware implementation, we utilized Vivado 2020.1. As per the LogicNets paper, we targeted the $\texttt{xcvu9p-flgb2104-2-i}$ FPGA part. The compilation was done using the \texttt{Flow\_PerfOptimized\_high} settings and configured in \texttt{Out-of-Context} (OOC) mode for synthesis. We aimed for a clock period of $1.6$ ns for the smaller networks and $2$ ns for the larger ones. The netlists' accuracy was verified using the LogicNets' pipeline, as the hardware architecture remained unchanged.

\subsection{Case study - Jet substructure tagging}
\label{section:case}
To evaluate the impact of polynomial degree on various performance metrics, we conducted a case study focusing on the jet substructure tagging task. This analysis focuses on how different degrees impact training loss, test accuracy, latency, and area. For this section, we used a fixed random sparsity mask and did not employ hardware-aware structured pruning techniques.

The primary goal is to show that increasing function complexity via polynomial expansion enhances training performance and enables ultra-low latency FPGA implementations. By concentrating on the jet substructure tagging task, we can gain valuable insights into the broader advantages and potential applications of our approach.

We adopted the JSC-M architecture from LogicNets as our baseline, detailed in Table~\ref{table:networks}. This architecture consists of five layers with parameters set to $\beta=3$ and $F=4$.

For our experiments, we trained the JSC-M network using polynomial degrees ranging from $1$ to $6$. Additionally, we systematically removed one hidden layer at a time, creating a set of sub-networks to evaluate the same range of polynomial degrees on different-sized neural network models. Each layer in the network is capable of being processed within a single clock cycle, and as a result we examined networks with inference speeds ranging from $2$ to $5$ clock cycles.

This methodology enabled us to comprehensively analyze how the removal of layers affects performance metrics such as training loss, test accuracy, latency, and area. By doing so, we could investigate the trade-off between inference speed and test accuracy, providing an understanding of how various polynomial configurations impact both accuracy and efficiency.

\begin{table*}[htbp]
\caption{Model architectures used as benchmarks for evaluated datasets.}
\begin{center}
\renewcommand{\arraystretch}{1.5} 
\begin{tabular}{cccccc}
\Xhline{2\arrayrulewidth}
\textbf{Dataset}&\textbf{Model Name}&\textbf{Nodes per Layer}&$\boldsymbol{\beta}$&\textbf{$F$}&\textbf{Exceptions}\\
\Xhline{2\arrayrulewidth}
Jet substructure & JSC-M & 64, 32, 32, 32, 5 & 3 & 4& \\
\hline
Jet substructure & JSC-M Lite & 64, 32, 5 & 3 & 4& \\
\hline
Jet substructure & JSC-XL & 128, 64, 64, 64, 5 & 5 & 3& $\beta_0=7$, $F_0=2$\\
\hline
UNSW-NB15 & NID Lite & 686, 147, 98, 49, 1 & 2 & 7&$\beta_0=1$\\
\hline
MNIST & HDR & 256, 100, 100, 100, 100, 10 & 2 & 6& \\
\hline

\end{tabular}
\renewcommand{\arraystretch}{1.2} 
\label{table:networks}
\end{center}
\end{table*}

\subsubsection{The impact of degree on training loss and test accuracy}
\begin{figure}[tbp]
	\centering
        \hspace{-1.1cm}
        \resizebox{1.1\columnwidth}{!}{\definecolor{myblue}{rgb}{0.35, 0.45, 0.64}
\definecolor{myorange}{rgb}{0.79, 0.53, 0.38}
\definecolor{myred}{rgb}{0.7, 0.36, 0.37}
\definecolor{mygreen}{rgb}{0.37, 0.61, 0.42}
\definecolor{mymagenta}{rgb}{0.52, 0.47, 0.66}
\definecolor{myyellow}{rgb}{1.0, 0.75, 0.0}
\definecolor{mycyan}{rgb}{0.0, 0.72, 0.92}
\definecolor{codegreen}{rgb}{0,0.6,0}

\begin{tikzpicture}
\pgfplotsset{compat=1.5}
\begin{axis}[
  width=1\columnwidth,
  height=50mm,
  grid=both,
  xtick={2,3,4,5},
  legend columns=-1,
  minor x tick num=1,
  minor y tick num=1,
  xlabel=Number of Layers,
  ylabel=Training Loss,
  tick label style={font=\footnotesize},  
  legend style={at={(0.5,1.2)},anchor=north, font=\footnotesize},
  every node/.style={
    font=\sffamily\scriptsize
    },  
    circtext/.style={draw,circle,minimum size=8pt,inner sep=2pt},
    dot/.style={draw,circle,fill=black,minimum size=0.6mm,inner sep=0pt},
]

\addlegendimage{empty legend}

\addplot [myyellow, smooth, thick, mark color=myblue,
mark=x]coordinates {
    (02, 0.8688)
    (03, 0.8081)
    (04, 0.793)
    (05, 0.7861)
};

\addplot [myorange,smooth, thick, mark color=myorange,
mark=x] coordinates {
    (02, 0.8169)
    (03, 0.7863)
    (04, 0.7752)
    (05, 0.7738)
};
\addplot [myred,smooth,thick,mark color=myred,
mark=x] 
coordinates {
    (02, 0.797)
    (03, 0.7803)
    (04, 0.7721)
    (05, 0.7774)
};
\addplot [mygreen,smooth,thick,mark color=mygreen,
mark=x] 
coordinates {
    (02, 0.8102)
    (03, 0.7779)
    (04, 0.7717)
    (05, 0.769)
};
\addplot [myblue,smooth,thick,mark color=myyellow,
mark=x] 
coordinates {
    (02, 0.8062)
    (03, 0.7908)
    (04, 0.7692)
    (05, 0.7692)
};
\addplot [mycyan,smooth,thick,mark color=mycyan,
mark=x] 
coordinates {
    (02, 0.8021)
    (03, 0.7749)
    (04, 0.7699)
    (05, 0.7668)
};

\draw[black,very thick,<-] (50,72) -- (110,90) node [fill=gray!30!white,pos=1.5, rounded corners=2pt, inner sep=2pt]
{Traditional Network};

\draw[black,very thick,<-] (210,11) -- (230,40) node [fill=gray!30!white,pos=1.3, rounded corners=2pt, inner sep=2pt]
{Our Networks};

\addlegendentry{\hspace{-.0cm}\textbf{Degree}}\addlegendentry{1}
\addlegendentry{2}
\addlegendentry{3}
\addlegendentry{4}
\addlegendentry{5}
\addlegendentry{6}
\end{axis}
\end{tikzpicture}}
	\caption{
            Training loss of models with varying numbers of layers across six different polynomial degrees.
        }
	\label{fig:g0}
\end{figure}
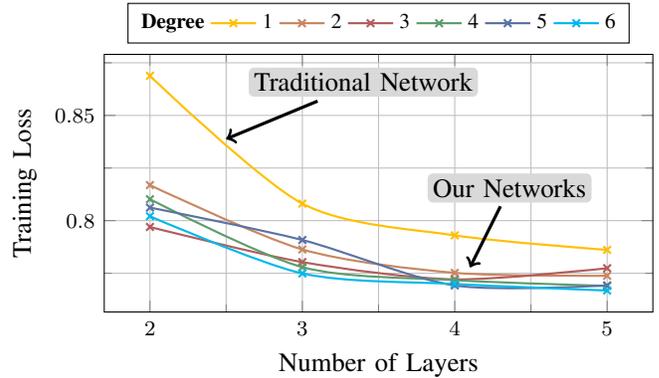

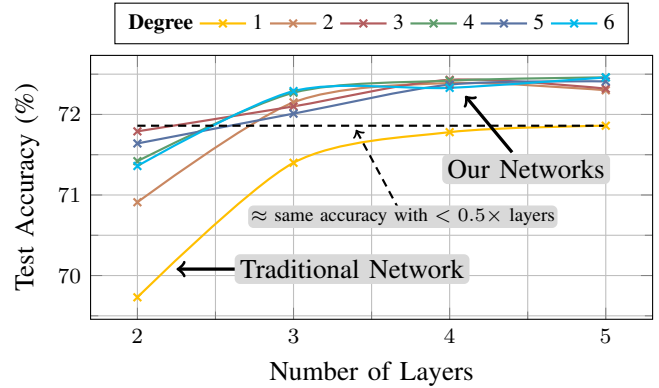
\begin{figure}[tbp]
	\centering
 \hspace{-1.1cm}
        \resizebox{1.1\columnwidth}{!}{\definecolor{myblue}{rgb}{0.35, 0.45, 0.64}
\definecolor{myorange}{rgb}{0.79, 0.53, 0.38}
\definecolor{myred}{rgb}{0.7, 0.36, 0.37}
\definecolor{mygreen}{rgb}{0.37, 0.61, 0.42}
\definecolor{mymagenta}{rgb}{0.52, 0.47, 0.66}
\definecolor{myyellow}{rgb}{1.0, 0.75, 0.0}
\definecolor{mycyan}{rgb}{0.0, 0.72, 0.92}
\definecolor{codegreen}{rgb}{0,0.6,0}

\begin{tikzpicture}
\pgfplotsset{compat=1.5}
\begin{axis}[
  width=1\columnwidth,
  height=50mm,
  grid=both,
  xtick={2,3,4,5},
  legend columns=-1,
  minor x tick num=1,
  minor y tick num=1,
  xlabel=Number of Layers,
  ylabel=Test Accuracy (\%),
  tick label style={font=\footnotesize},  
  legend style={at={(0.5,1.2)},anchor=north, font=\footnotesize},
]
\addlegendimage{empty legend}

\addplot [myyellow,smooth, thick, mark color=myblue,
mark=x]coordinates {
    (02, 69.73)
    (03, 71.4)
    (04, 71.78)
    (05, 71.86)
};
\addplot [myorange,smooth, thick, mark color=myorange,
mark=x] coordinates {
    (02, 70.91)
    (03, 72.15)
    (04, 72.39)
    (05, 72.3)
};
\addplot [myred,smooth,thick,mark color=myred,
mark=x] 
coordinates {
    (02, 71.79)
    (03, 72.1)
    (04, 72.43)
    (05, 72.32)
};
\addplot [mygreen,smooth,thick,mark color=mygreen,
mark=x] 
coordinates {
    (02, 71.42)
    (03, 72.27)
    (04, 72.42)
    (05, 72.46)
};
\addplot [myblue,smooth,thick,mark color=myyellow,
mark=x] 
coordinates {
    (02, 71.64)
    (03, 72.01)
    (04, 72.37)
    (05, 72.41)
};
\addplot [mycyan,smooth,thick,mark color=mycyan,
mark=x] 
coordinates {
    (02, 71.36)
    (03, 72.29)
    (04, 72.33)
    (05, 72.46)
};

\addplot[densely dashed, black, thick]
coordinates {(02, 71.86) ((05, 71.86)};

\draw[black,very thick,<-] (25,35) -- (100,35) node [fill=gray!30!white,pos=1.5, rounded corners=2pt, inner sep=2pt]
{Traditional Network};

\draw[black,thick,densely dashed, <-] (140,208) -- (170,100) node[fill=gray!30, text=black, font=\scriptsize, rounded corners=2pt, inner sep=1pt] {$\approx$ same accuracy with $< 0.5\times$ layers};;

\draw[black,very thick,<-] (210,250) -- (240,180) node [fill=gray!30!white,pos=1.3, rounded corners=2pt, inner sep=2pt]
{Our Networks};

\addlegendentry{\hspace{-.0cm}\textbf{Degree}}
\addlegendentry{1}
\addlegendentry{2}
\addlegendentry{3}
\addlegendentry{4}
\addlegendentry{5}
\addlegendentry{6}
\end{axis}
\end{tikzpicture}}
	\caption{
            Test accuracy of models with varying number of layers across six different polynomial degrees.
        }
	\label{fig:g1}
\end{figure}

Firstly, we analyze how gradually increasing the polynomial degree affects both training loss and test accuracy. The results are depicted in Figure~\ref{fig:g0} and Figure~\ref{fig:g1}. 

The trends in training loss reveal that the curve for the models with $D=1$ is consistently above those for higher degrees, indicating that models with $D=1$ are less effective at minimizing loss. As the degree increases from $1$ to $2$, there is a noticeable improvement in minimizing training loss. However, beyond degree $2$, the improvements become less consistent, suggesting diminishing returns with very high degrees. This is likely because higher degrees increase function complexity, which can better fit the data but also pose greater challenges during training.

A similar trend is observed in test accuracy. Higher degrees initially improve accuracy, but the separation between data points becomes more unpredictable with very high degrees, indicating potential overfitting. High-degree models may also be more challenging to train, suggesting that fine-grained optimization may be required.

The most valuable observation from these graphs is that polynomial function training can achieve performance boosts equivalent to adding more layers to a linear network. Specifically, a $5$-layer linear network shows lower accuracy compared to a $3$-layer polynomial network, and almost matches the accuracy of a $2$-layer polynomial network. This demonstrates the effectiveness of multivariate polynomial function training in enhancing network performance, allowing for comparable or even superior accuracy with fewer layers.

These findings underscore the advantages of using polynomial functions, as they provide the flexibility to achieve efficient models without the need for very deep networks. This flexibility can compensate for the complexity associated with training deeper networks, resulting in more efficient and potentially more accurate models.

\subsubsection{The impact of degree on latency and resource utilization}
Previously, we explored PolyLUT networks can maintain the same accuracy with fewer layers. This layer reduction directly translates to a decrease in inference speed, with each removed layer reducing latency by one clock cycle.

\begin{figure}[tbp]
	\centering
 \hspace{-1.1cm}
        \resizebox{1.1\columnwidth}{!}{\definecolor{myblue}{rgb}{0.35, 0.45, 0.64}
\definecolor{myorange}{rgb}{0.79, 0.53, 0.38}
\definecolor{myred}{rgb}{0.7, 0.36, 0.37}
\definecolor{mygreen}{rgb}{0.37, 0.61, 0.42}
\definecolor{mymagenta}{rgb}{0.52, 0.47, 0.66}
\definecolor{myyellow}{rgb}{1.0, 0.75, 0.0}
\definecolor{mycyan}{rgb}{0.0, 0.72, 0.92}
\definecolor{codegreen}{rgb}{0,0.6,0}

\begin{tikzpicture}
\pgfplotsset{compat=1.5}
\begin{axis}[scatter/classes={
  a={mark=*,myyellow, scale=0.8},
  b={mark=*,myorange, scale=0.8},
  c={mark=*,myred, scale=0.8},
  d={mark=*,mygreen, scale=0.8},
  e={mark=*,myblue, scale=0.8},
  f={mark=*,mycyan, scale=0.8}},
  width=1\columnwidth,
  height=50mm,
  grid=both,
  legend columns=-1,
  minor x tick num=1,
  minor y tick num=1,
  set layers, 
  mark layer=axis tick labels,
  xlabel=Latency (ns),
  ylabel=Test Error Rate (\%),
  tick label style={font=\footnotesize},  
  legend style={at={(0.5,1.2)},anchor=north, font=\footnotesize},
]
\addlegendimage{empty legend}

\addplot [scatter, only marks, scatter src=explicit symbolic]coordinates {
    (2.476, 30.27) [a]
    (4.797, 28.6) [a]
    (6.38, 28.22) [a]
    (8.9, 28.14) [a]
};
\addplot [scatter, only marks, scatter src=explicit symbolic] coordinates {
    (2.624, 29.09) [b]
    (4.659, 27.85) [b]
    (7.456, 27.61) [b]
    (11.285, 27.7) [b]
};
\addplot [scatter, only marks, scatter src=explicit symbolic] 
coordinates {
    (2.784, 28.21) [c]
    (4.884, 27.9) [c]
    (6.692, 27.57) [c]
    (11.63, 27.68) [c]
};
\addplot [scatter, only marks, scatter src=explicit symbolic] 
coordinates {
    (3.08, 28.58) [d]
    (4.854, 27.73) [d]
    (7.092, 27.58) [d]
    (9.755, 27.54) [d]
};
\addplot [scatter, only marks, scatter src=explicit symbolic] 
coordinates {
    (2.724, 28.36) [e]
    (4.716, 27.99) [e]
    (7.644, 27.63) [e]
    (10.44, 27.59) [e]
};
\addplot [scatter, only marks, scatter src=explicit symbolic] 
coordinates {
    (2.684, 28.64) [f]
    (4.647, 27.71) [f]
    (7.248, 27.67) [f]
    (9.4, 27.54) [f]
};
\addplot [black,thick] 
coordinates {
    (2.476, 30.27)
    (4.797, 30.27)
    (4.797, 28.6)
    (6.38, 28.6)
    (6.38, 28.22)
    (8.9, 28.22)
    (8.9, 28.14)
};

\addplot [red,thick] 
coordinates {
    (2.476, 30.27)
    (2.624, 30.27)
    (2.624, 29.09)
    (2.684, 29.09)
    (2.684, 28.64)
    (2.724, 28.64)
    (2.724, 28.35)
    (2.784, 28.35)
    (2.784, 28.19)
    (4.62, 28.19)
    (4.62, 27.69)
    (6.692, 27.69)
    (6.692, 27.57)
    (9.4, 27.57)
    (9.4, 27.54)
};

\addplot [black, densely dashed, smooth,<-]
coordinates {
    (4.797, 28.6)
    (3.7605, 28.73)
    (2.724, 28.6)
}
node[
  text=black,
  font=\tiny,
  above,
  yshift=-0.1cm] at (1215,28.6) [pos=0.5,font=\tiny]{$\mathbf{1.76\times}$};;

\addplot [black, densely dashed, smooth,<-]
coordinates {
    (6.38, 28.22)
    (4.582, 28.32)
    (2.784, 28.22)
}
node[
  text=black,
  font=\tiny,
  above,
  yshift=-0.1cm] at (1215,28.6) [pos=0.5,font=\tiny]{$\mathbf{2.29\times}$};;

\addplot [black, densely dashed,<-]
coordinates {
    (8.9, 28.14)
    (6.76, 28.04)
    (4.62, 28.14)
}
node[
  text=black,
  font=\tiny,
  below,
  yshift=0.08cm] at (1215,28.6) [pos=0.5,font=\tiny]{$\mathbf{1.92\times}$};;

\draw[black,very thick,<-] (250,230) -- (530,230) node [fill=gray!30!white,pos=1.5, rounded corners=2pt, inner sep=2pt]
{Baseline Pareto Frontier};

\draw[black,very thick,<-] (640,10) -- (710,90) node [fill=gray!30!white,pos=1.3, rounded corners=2pt, inner sep=2pt]
{Our Pareto Frontier};

\addlegendentry{\hspace{-.0cm}\textbf{Degree}}
\addlegendentry{1}
\addlegendentry{2}
\addlegendentry{3}
\addlegendentry{4}
\addlegendentry{5}
\addlegendentry{6}
\end{axis}
\end{tikzpicture}}
	\caption{
            Trade-off analysis between test error rate and latency for models with varying numbers of layers across six different polynomial degrees. Our improved Pareto frontier is depicted by (\ref{pgfplots:label1}).
        }
	\label{fig:g3}
\end{figure}
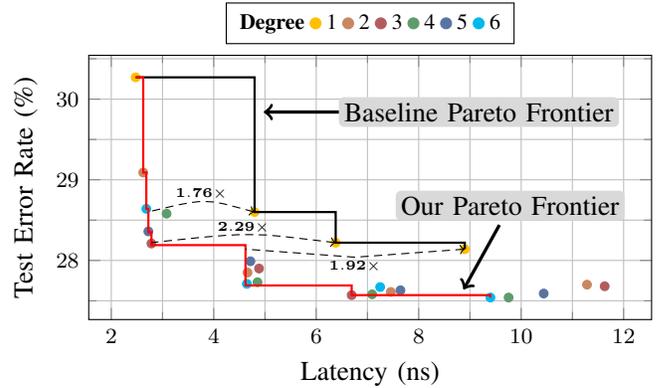

Figure~\ref{fig:g3} shows the latency and test error rates for the same models discussed in the previous section. The Pareto frontier for polynomial configurations demonstrates a more efficient trade-off between latency and test error rate compared to the linear configurations. Specifically, for the same test error rate, latency reductions range from $1.76\times$ to $2.29\times$.

While the theoretical LUT cost would be the same regardless of the polynomial degree, practical implementations reveal differences. This is because, in practice, L-LUTs with $N$ inputs will naturally synthesize to smaller circuits than others, when implemented within a network of native FPGA P-LUTs with $K$ inputs, when $N > K$. This can be seen in Figure~\ref{fig:g3} which illustrates the LUT count of all data points from the previous section alongside their respective test error rates. 

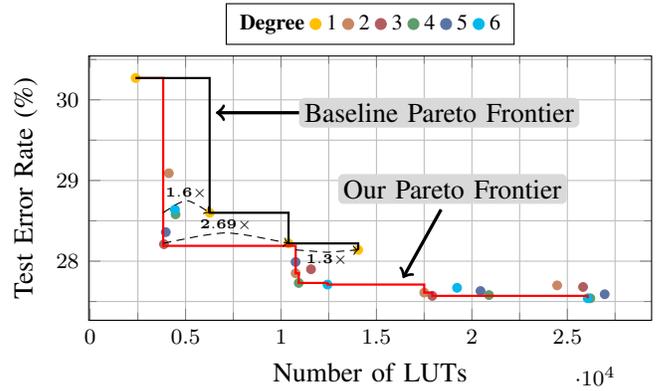
\begin{figure}[tbp]
	\centering
 \hspace{-1.1cm}
        \resizebox{1.1\columnwidth}{!}{\definecolor{myblue}{rgb}{0.35, 0.45, 0.64}
\definecolor{myorange}{rgb}{0.79, 0.53, 0.38}
\definecolor{myred}{rgb}{0.7, 0.36, 0.37}
\definecolor{mygreen}{rgb}{0.37, 0.61, 0.42}
\definecolor{mymagenta}{rgb}{0.52, 0.47, 0.66}
\definecolor{myyellow}{rgb}{1.0, 0.75, 0.0}
\definecolor{mycyan}{rgb}{0.0, 0.72, 0.92}
\definecolor{codegreen}{rgb}{0,0.6,0}

\begin{tikzpicture}
\pgfplotsset{compat=1.5}
\begin{axis}[scatter/classes={
  a={mark=*,myyellow, scale=0.8},
  b={mark=*,myorange, scale=0.8},
  c={mark=*,myred, scale=0.8},
  d={mark=*,mygreen, scale=0.8},
  e={mark=*,myblue, scale=0.8},
  f={mark=*,mycyan, scale=0.8}},
  width=1\columnwidth,
  height=50mm,
  grid=both,
  set layers, 
  mark layer=axis tick labels,
  legend columns=-1,
  minor x tick num=1,
  minor y tick num=1,
  xlabel=Number of LUTs,
  ylabel=Test Error Rate (\%),
  tick label style={font=\footnotesize},  
  legend style={at={(0.5,1.2)},anchor=north, font=\footnotesize},
]
\addlegendimage{empty legend}

\addplot [scatter, only marks, scatter src=explicit symbolic]coordinates {
    (2385, 30.27) [a]
    (6261, 28.6) [a]
    (10394, 28.22) [a]
    (14044, 28.14) [a]
};
\addplot [scatter, only marks, scatter src=explicit symbolic] coordinates {
    (4124, 29.09) [b]
    (10768, 27.85) [b]
    (17502, 27.61) [b]
    (24465, 27.7) [b]
};
\addplot [scatter, only marks, scatter src=explicit symbolic] 
coordinates {
    (3870, 28.21) [c]
    (11572, 27.9) [c]
    (17923, 27.57) [c]
    (25822, 27.68) [c]
};
\addplot [scatter, only marks, scatter src=explicit symbolic] 
coordinates {
    (4476, 28.58) [d]
    (10929, 27.73) [d]
    (20894, 27.58) [d]
    (26201, 27.54) [d]
};
\addplot [scatter, only marks, scatter src=explicit symbolic] 
coordinates {
    (3965, 28.36) [e]
    (10764, 27.99) [e]
    (20449, 27.63) [e]
    (26955, 27.59) [e]
};
\addplot [scatter, only marks, scatter src=explicit symbolic] 
coordinates {
    (4442, 28.64) [f]
    (12436, 27.71) [f]
    (19221, 27.67) [f]
    (26066, 27.54) [f]
};

\addplot [red,thick] 
coordinates {
    (2385, 30.27)
    (3830, 30.27)
    (3830, 28.19)
    (10768, 28.19)
    (10768, 27.85)
    (10929, 27.85)
    (10929, 27.73)
    (12436, 27.73)
    (12436, 27.71)
    (17502, 27.71)
    (17502, 27.61)
    (17923, 27.61)
    (17923, 27.57)
    (26066, 27.57)
    (26066, 27.54)
};
\label{pgfplots:label1}
\addplot [black,thick] 
coordinates {
    (2385, 30.27)
    (6261, 30.27)
    (6261, 28.6)
    (10394, 28.6)
    (10394, 28.22)
    (14044, 28.22)
    (14044, 28.14)
};
\label{pgfplots:label2}

\addplot [black, densely dashed, smooth,->]
coordinates {
    (3830, 28.6)
    (5045, 28.75)
    (6261, 28.6)
}
node[
  text=black,
  font=\tiny,
  above,
  yshift=-0.1cm] at (1215,28.6) [pos=0.5,font=\tiny]{$\mathbf{1.6\times}$};;

\addplot [black, densely dashed, smooth,->]
coordinates {
    (3830, 28.22)
    (7112, 28.35)
    (10394, 28.22)
}
node[
  text=black,
  font=\tiny,
  above,
  yshift=-0.1cm] at (1215,28.6) [pos=0.5,font=\tiny]{$\mathbf{2.69\times}$};;

\addplot [black, densely dashed,->]
coordinates {
    (10768, 28.14)
    (12406, 28.11)
    (14044, 28.14)
}
node[
  text=black,
  font=\tiny,
  below,
  yshift=+0.1cm] at (1215,28.6) [pos=0.5,font=\tiny]{$\mathbf{1.3\times}$};;
 
\draw[black,very thick,<-] (42,230) -- (120,230) node [fill=gray!30!white,pos=1.5, rounded corners=2pt, inner sep=2pt]
{Baseline Pareto Frontier};

\draw[black,very thick,<-] (140,25) -- (160,110) node [fill=gray!30!white,pos=1.3, rounded corners=2pt, inner sep=2pt]
{Our Pareto Frontier};

\addlegendentry{\hspace{-.0cm}\textbf{Degree}}
\addlegendentry{1}
\addlegendentry{2}
\addlegendentry{3}
\addlegendentry{4}
\addlegendentry{5}
\addlegendentry{6}
\end{axis}
\end{tikzpicture}}
	\caption{
            Trade-off analysis between test error rate and resource utilization for models with varying numbers of layers across six different polynomial degrees. Our improved Pareto frontier is depicted by (\ref{pgfplots:label1}).
        }
	\label{fig:g2}
\end{figure}

As the polynomial degree increases, slightly more P-LUTs are required for the same network topology due to the enhanced complexity of the functions being expressed. However, the increase in LUT utilization from higher degrees is completely offset by the reduction in the number of layers. This overall leads to savings in the number of LUTs, as shown in  Figure~\ref{fig:g2}, where polynomial-trained designs form a more efficient Pareto frontier than linear networks.

Comparing a $2$-layer network with $D=3$ (which yields an accuracy of $71.79\%$) to a $4$-layer network with $D=1$ (which yields a comparable accuracy of $71.78\%$), we observe a $2.29\times$ improvement in latency and a $2.69\times$ improvement in resource utilization. This highlights the benefits of implementing multivariate polynomial functions within the LUTs, effectively reducing both inference latency and resource utilization.

\subsection{Stochasticity Study}
\label{section:stochastic}
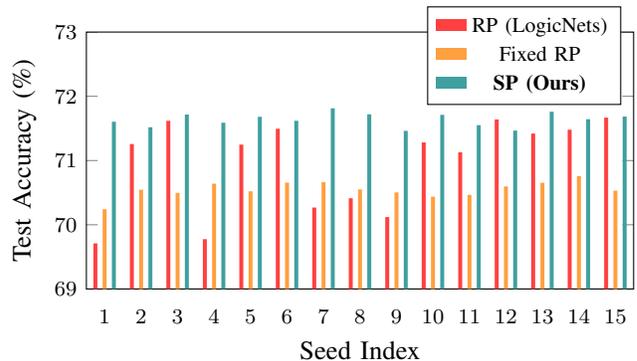
\begin{figure}
    \definecolor{lutcolor}{HTML}{b2df8a}
\definecolor{ffcolor}{HTML}{1f78b4}
\definecolor{fmaxcolor}{HTML}{a6cee3}
\definecolor{pgfred}{HTML}{e31a1c}

\begin{tikzpicture}
\pgfplotsset{set layers, compat=1.5}
\begin{axis}[
    ybar,
    bar width=1.5pt,
    height=50mm,
    width=\columnwidth,
    xlabel={Seed Index},
    ylabel={Test Accuracy (\%)},
    xtick={1,2,3,4,5,6,7,8,9,10,11,12,13,14,15},
    ymin=69, ymax=73,
    xmin=0.5, xmax=15.5,
    xtick style={draw=none},
    tick label style={font=\footnotesize},
    legend style={at={(0.8,1.1)},anchor=north, font=\footnotesize},
    legend image code/.code={%
        \draw[#1] (0cm,-0.05cm) rectangle (0.4cm,0.1cm);
    },
  ]
 
\addplot [fill=red!75, draw opacity=0] 
coordinates {
(1,69.710)
(2,71.257)
(3,71.619)
(4,69.776)
(5,71.248)
(6,71.495)
(7,70.268)
(8,70.413)
(9,70.120)
(10,71.282)
(11,71.128)
(12,71.640)
(13,71.420)
(14,71.48)
(15,71.6683)

}; 
\addlegendentry{\footnotesize RP (LogicNets)}

\addplot [fill=orange!75, draw opacity=0] 
coordinates {
(1,70.24249)
(2,70.5454)
(3,70.4968)
(4,70.63973)
(5,70.5211)
(6,70.65645)
(7,70.66253)
(8,70.54954)
(9,70.50394)
(10,70.43706)
(11,70.46543)
(12,70.59717)
(13,70.65341)
(14,70.7567)
(15,70.5308)
};
\addlegendentry{\footnotesize Fixed RP}

\addplot [fill=teal!75, draw opacity=0] 
coordinates {
(1,71.6044)
(2,71.5188)
(3,71.7159)
(4,71.5882)
(5,71.6809)
(6,71.6191)
(7,71.8132)
(8,71.7204)
(9,71.4615)
(10,71.7113)
(11,71.5512)
(12,71.4661)
(13,71.760)
(14,71.6439)
(15,71.6845)
};
\addlegendentry{\footnotesize \textbf{SP (Ours)}}

\end{axis}
\end{tikzpicture}
    \caption{
        The standard deviation of test accuracy between seeds is 0.679 for the default design with random pruning (RP), 0.118 for the default design with a fixed RP mask, and 0.101 for the designs with structured pruning (SP).
    }
    \label{plot:seeds}
\end{figure}

We analyzed the sources of stochasticity in our neural networks using the Jet Substructure dataset, as illustrated in Figure~\ref{plot:seeds}. The default setting employs the \textit{a priori} sparsity technique from LogicNets. The observed standard deviation in test accuracy between different random seeds is $0.679$, which is often deemed unacceptable, especially in neural architecture search. This process relies heavily on accuracy as a performance metric, and high volatility in this metric impedes the ability to correlate network characteristics with performance effectively.

By applying a fixed sparsity mask, the standard deviation drops to $0.118$. This reduction indicates that most of the stochasticity originates from the random mask rather than the optimization algorithms. Therefore, we propose using a hardware-aware regularizer to perform structured pruning (SP) instead of random pruning (RP) to eliminate the stochasticity caused by the random sparsity mask. The standard deviation for the SP algorithm is the lowest at $0.101$, and it also improves test accuracy, surpassing each individual result of the method with random sparsity.

\subsection{Combined evaluation - Jet substructure tagging}
\label{section:ablation1}
We have rerun the tests on the jet substructure tagging task from the case study, focusing on the performance advantage of combining the polynomial function methodology ($D=2$) with the hardware-aware structured pruning. By examining the test accuracy and the training loss across different seeds, we demonstrate a consistent improvement over the baseline and a significant reduction in variation across seeds, particularly in small models. It is important to note that the goal is to utilize as small as possible models, therefore improving the robustness and the performance of the small models would significantly reduce the area and latency.

By examining the trends in training loss in Figure~\ref{fig:a0}, we demonstrate that our proposed method of training piecewise polynomial functions with learned connectivity outperforms the LogicNets baseline (Degree 1 RP) across all seeds. Additionally, it consistently delivers high-performance designs regardless of the seed, even for extremely small models, such as networks with only 2 or 3 layers. 

The same advantages are evident in terms of test accuracy, as shown in Figure~\ref{fig:a1}. It can be observed that the Degree 2 SP design points have very small variations in test accuracy across seeds ($<0.25pp$), whereas the Degree 1 RP (LogicNets baseline) ones could have differences of $>1pp$, which may be unacceptable for many automated search algorithms. Consequently, a 2-layer Degree 2 SP design can successfully replace a 5-layer Degree 1 RP design, matching its performance both in terms of test accuracy and robustness.

\begin{figure}[tbp]
	\centering
        \hspace{-1.1cm}
        \resizebox{1.1\columnwidth}{!}{\definecolor{myblue}{rgb}{0.35, 0.45, 0.64}
\definecolor{myorange}{rgb}{0.79, 0.53, 0.38}
\definecolor{myred}{rgb}{0.7, 0.36, 0.37}
\definecolor{mygreen}{rgb}{0.37, 0.61, 0.42}
\definecolor{mymagenta}{rgb}{0.52, 0.47, 0.66}
\definecolor{myyellow}{rgb}{1.0, 0.75, 0.0}
\definecolor{mycyan}{rgb}{0.0, 0.72, 0.92}
\definecolor{codegreen}{rgb}{0,0.6,0}

\begin{tikzpicture}
\pgfplotsset{compat=1.5}
\begin{axis}[
  width=1\columnwidth,
  height=50mm,
  grid=both,
  xtick={2,3,4,5},
  legend columns=-1,
  minor x tick num=1,
  minor y tick num=1,
  xlabel=Number of Layers,
  ylabel=Training Loss,
  tick label style={font=\footnotesize},  
  legend style={at={(0.48,1.3)},anchor=north, font=\footnotesize},
]
\addlegendimage{empty legend}
\addplot [red,smooth,thick,mark color=red,
mark=x] 
coordinates {
    (02, 0.790286303)
    (03, 0.779619455)
    (04, 0.775357842)
    (05, 0.772862077)
};
\addplot [myblue,smooth,thick,mark color=myblue,
mark=x] 
coordinates {
    (02, 0.798042953)
    (03, 0.781444073)
    (04, 0.776402593)
    (05, 0.773492515)
};
\addplot [myyellow,smooth, thick, mark color=myyellow,
mark=x]coordinates {
    (02, 0.824062526)
    (03, 0.798363149)
    (04, 0.788713813)
    (05, 0.78469938)
};
\addplot [myyellow,only marks, thick, mark color=myyellow,
mark=x]coordinates {
    (02, 0.855748653)
    (03, 0.811317325)
    (04, 0.792799532)
    (05, 0.789538443)
};
\addplot [myyellow,only marks, thick, mark color=myyellow,
mark=x]coordinates {
    (02, 0.832701504)
    (03, 0.804740548)
    (04, 0.79179436)
    (05, 0.787285924)
};
\addplot [myyellow,only marks, thick, mark color=myyellow,
mark=x]coordinates {
    (02, 0.846109033)
    (03, 0.799717426)
    (04, 0.791098177)
    (05, 0.787275195)
};
\addplot [myyellow,only marks, thick, mark color=myyellow,
mark=x]coordinates {
    (02, 0.850024283)
    (03, 0.803612411)
    (04, 0.791161239)
    (05, 0.788371801)
};
\addplot [myyellow,only marks, thick, mark color=myyellow,
mark=x]coordinates {
    (02, 0.83351028)
    (03, 0.81282258)
    (04, 0.793735683)
    (05, 0.789604247)
};
\addplot [myyellow,only marks, thick, mark color=myyellow,
mark=x]coordinates {
    (02, 0.870313406)
    (03, 0.808597684)
    (04, 0.793087006)
    (05, 0.786787748)
};
\addplot [myyellow,only marks, thick, mark color=myyellow,
mark=x]coordinates {
    (02, 0.849852443)
    (03, 0.803453386)
    (04, 0.793434799)
    (05, 0.790555716)
};
\addplot [myyellow,only marks, thick, mark color=myyellow,
mark=x]coordinates {
    (02, 0.858069837)
    (03, 0.805838346)
    (04, 0.792803943)
    (05, 0.789509416)
};
\addplot [myyellow,only marks, thick, mark color=myyellow,
mark=x]coordinates {
    (02, 0.900611341)
    (03, 0.808384061)
    (04, 0.792198956)
    (05, 0.788280547)
};

\addplot [myblue,only marks, thick, mark color=myblue,
mark=x]coordinates {
    (02, 0.816745937)
    (03, 0.785589218)
    (04, 0.778608024)
    (05, 0.777090669)
};
\addplot [myblue,only marks, thick, mark color=myblue,
mark=x]coordinates {
    (02, 0.806256831)
    (03, 0.794107318)
    (04, 0.77756834)
    (05, 0.775022626)
};
\addplot [myblue,only marks, thick, mark color=myblue,
mark=x]coordinates {
    (02, 0.813848376)
    (03, 0.783848763)
    (04, 0.78130281)
    (05, 0.774456799)
};
\addplot [myblue,only marks, thick, mark color=myblue,
mark=x]coordinates {
    (02, 0.810993433)
    (03, 0.785417795)
    (04, 0.776656508)
    (05, 0.775455296)
};
\addplot [myblue,only marks, thick, mark color=myblue,
mark=x]coordinates {
    (02, 0.805636942)
    (03, 0.787798703)
    (04, 0.782973468)
    (05, 0.7754426)
};
\addplot [myblue,only marks, thick, mark color=myblue,
mark=x]coordinates {
    (02, 0.818608105)
    (03, 0.787858069)
    (04, 0.777680874)
    (05, 0.775639176)
};
\addplot [myblue,only marks, thick, mark color=myblue,
mark=x]coordinates {
    (02, 0.815690696)
    (03, 0.787089467)
    (04, 0.778908968)
    (05, 0.777280271)
};
\addplot [myblue,only marks, thick, mark color=myblue,
mark=x]coordinates {
    (02, 0.826061606)
    (03, 0.786355138)
    (04, 0.777781487)
    (05, 0.775440335)
};
\addplot [myblue,only marks, thick, mark color=myblue,
mark=x]coordinates {
    (02, 0.864291787)
    (03, 0.784629583)
    (04, 0.782064676)
    (05, 0.776244998)
};

\addplot [red,only marks, thick, mark color=red,
mark=x]coordinates {
    (02, 0.792261541)
    (03, 0.78114599)
    (04, 0.776977003)
    (05, 0.774576724)
};
\addplot [red,only marks, thick, mark color=red,
mark=x]coordinates {
    (02, 0.794801414)
    (03, 0.783458471)
    (04, 0.776961386)
    (05, 0.774416149)
};
\addplot [red,only marks, thick, mark color=red,
mark=x]coordinates {
    (02, 0.796607971)
    (03, 0.785289466)
    (04, 0.774694204)
    (05, 0.775364518)
};
\addplot [red,only marks, thick, mark color=red,
mark=x]coordinates {
    (02, 0.792867064)
    (03, 0.781432092)
    (04, 0.775945067)
    (05, 0.774119675)
};
\addplot [red,only marks, thick, mark color=red,
mark=x]coordinates {
    (02, 0.790606141)
    (03, 0.780826747)
    (04, 0.77761507)
    (05, 0.775911689)
};
\addplot [red,only marks, thick, mark color=red,
mark=x]coordinates {
    (02, 0.791380405)
    (03, 0.781732798)
    (04, 0.776076496)
    (05, 0.775280535)
};
\addplot [red,only marks, thick, mark color=red,
mark=x]coordinates {
    (02, 0.793373168)
    (03, 0.781679571)
    (04, 0.776743591)
    (05, 0.773523629)
};
\addplot [red,only marks, thick, mark color=red,
mark=x]coordinates {
    (02, 0.798592269)
    (03, 0.782120228)
    (04, 0.775450587)
    (05, 0.773881137)
};
\addplot [red,only marks, thick, mark color=red,
mark=x]coordinates {
    (02, 0.797124624)
    (03, 0.780671537)
    (04, 0.776434302)
    (05, 0.774208248)
};

\addlegendentry{\hspace{-.0cm}\textbf{Degree}}
\addlegendentry{Degree 2 SP}
\addlegendentry{Degree 2 RP}
\addlegendentry{Degree 1 RP}

\end{axis}
\end{tikzpicture}}
	\caption{
            Training loss of models with varying numbers of layers across ten different seeds.
        }
	\label{fig:a0}
\end{figure}

\begin{figure}[tbp]
	\centering
 \hspace{-1.1cm}
        \resizebox{1.1\columnwidth}{!}{\definecolor{myblue}{rgb}{0.35, 0.45, 0.64}
\definecolor{myorange}{rgb}{0.79, 0.53, 0.38}
\definecolor{myred}{rgb}{0.7, 0.36, 0.37}
\definecolor{mygreen}{rgb}{0.37, 0.61, 0.42}
\definecolor{mymagenta}{rgb}{0.52, 0.47, 0.66}
\definecolor{myyellow}{rgb}{1.0, 0.75, 0.0}
\definecolor{mycyan}{rgb}{0.0, 0.72, 0.92}
\definecolor{codegreen}{rgb}{0,0.6,0}

\begin{tikzpicture}
\pgfplotsset{compat=1.5}
\begin{axis}[
  width=1\columnwidth,
  height=50mm,
  grid=both,
  xtick={2,3,4,5},
  legend columns=-1,
  minor x tick num=1,
  minor y tick num=1,
  xlabel=Number of Layers,
  ylabel=Test Accuracy (\%),
  tick label style={font=\footnotesize},  
  legend style={at={(0.48,1.3)},anchor=north, font=\footnotesize},
]
\addlegendimage{empty legend}
\addplot [red,smooth,thick,mark color=red,
mark=x] 
coordinates {
    (02, 71.99)
    (03, 72.2)
    (04, 72.32)
    (05, 72.42)
};
\addplot [myblue,smooth,thick,mark color=myblue,
mark=x] 
coordinates {
    (02, 71.66)
    (03, 72.029)
    (04, 72.25)
    (05, 72.29)
};
\addplot [myyellow,smooth, thick, mark color=myyellow,
mark=x]coordinates {
    (02, 71.31)
    (03, 71.83)
    (04, 71.97)
    (05, 72.087)
};
\addplot [myyellow,only marks, thick, mark color=myyellow,
mark=x]coordinates {
    (02, 69.923)
    (03, 71.41)
    (04, 71.86)
    (05, 71.88)
};
\addplot [myyellow,only marks, thick, mark color=myyellow,
mark=x]coordinates {
    (02, 71.20)
    (03, 71.79)
    (04, 71.94)
    (05, 71.93)
};
\addplot [myyellow,only marks, thick, mark color=myyellow,
mark=x]coordinates {
    (02, 71.04)
    (03, 71.77)
    (04, 71.893)
    (05, 72.02)
};
\addplot [myyellow,only marks, thick, mark color=myyellow,
mark=x]coordinates {
    (02, 71.01)
    (03, 71.64)
    (04, 71.891)
    (05, 71.94)
};
\addplot [myyellow,only marks, thick, mark color=myyellow,
mark=x]coordinates {
    (02, 71.21)
    (03, 71.54)
    (04, 71.84)
    (05, 72.071)
};
\addplot [myyellow,only marks, thick, mark color=myyellow,
mark=x]coordinates {
    (02, 70.65)
    (03, 71.53)
    (04, 71.96)
    (05, 72.085)
};
\addplot [myyellow,only marks, thick, mark color=myyellow,
mark=x]coordinates {
    (02, 70.55)
    (03, 71.8)
    (04, 71.9)
    (05, 71.95)
};
\addplot [myyellow,only marks, thick, mark color=myyellow,
mark=x]coordinates {
    (02, 69.923)
    (03, 71.58)
    (04, 71.96)
    (05, 71.87)
};
\addplot [myyellow,only marks, thick, mark color=myyellow,
mark=x]coordinates {
    (02, 68.53)
    (03, 71.44)
    (04, 71.83)
    (05, 72.077)
};

\addplot [myblue,only marks,thick,mark color=myblue,
mark=x] 
coordinates {
    (02, 71.52)
    (03, 71.993)
    (04, 72.127)
    (05, 72.274)
};
\addplot [myblue,only marks,thick,mark color=myblue,
mark=x] 
coordinates {
    (02, 71.42)
    (03, 71.91)
    (04, 72.13)
    (05, 72.229)
};
\addplot [myblue,only marks,thick,mark color=myblue,
mark=x] 
coordinates {
    (02, 71.65)
    (03, 72.023)
    (04, 72.17)
    (05, 72.257)
};
\addplot [myblue,only marks,thick,mark color=myblue,
mark=x] 
coordinates {
    (02, 71.14)
    (03, 71.99)
    (04, 72.123)
    (05, 72.197)
};
\addplot [myblue,only marks,thick,mark color=myblue,
mark=x] 
coordinates {
    (02, 71.44)
    (03, 71.73)
    (04, 72.07)
    (05, 72.276)
};
\addplot [myblue,only marks,thick,mark color=myblue,
mark=x] 
coordinates {
    (02, 71.65)
    (03, 72.024)
    (04, 72.14)
    (05, 72.253)
};
\addplot [myblue,only marks,thick,mark color=myblue,
mark=x] 
coordinates {
    (02, 71.61)
    (03, 72.01)
    (04, 72.23)
    (05, 72.143)
};
\addplot [myblue,only marks,thick,mark color=myblue,
mark=x] 
coordinates {
    (02, 70.84)
    (03, 71.89)
    (04, 72.18)
    (05, 72.26)
};
\addplot [myblue,only marks,thick,mark color=myblue,
mark=x] 
coordinates {
    (02, 69.34)
    (03, 72.027)
    (04, 72.077)
    (05, 72.253)
};

\addplot [red,only marks,thick,mark color=red,
mark=x] 
coordinates {
    (02, 71.89)
    (03, 72.05)
    (04, 72.12)
    (05, 72.24)
};
\addplot [red,only marks,thick,mark color=red,
mark=x] 
coordinates {
    (02, 71.66)
    (03, 72.08)
    (04, 72.18)
    (05, 72.27)
};
\addplot [red,only marks,thick,mark color=red,
mark=x] 
coordinates {
    (02, 71.87)
    (03, 72.1)
    (04, 72.17)
    (05, 72.32)
};
\addplot [red,only marks,thick,mark color=red,
mark=x] 
coordinates {
    (02, 71.85)
    (03, 72.0)
    (04, 72.07)
    (05, 72.22)
};
\addplot [red,only marks,thick,mark color=red,
mark=x] 
coordinates {
    (02, 71.98)
    (03, 72)
    (04, 72.17)
    (05, 72.332)
};
\addplot [red,only marks,thick,mark color=red,
mark=x] 
coordinates {
    (02, 71.845)
    (03, 72.09)
    (04, 72.26)
    (05, 72.33)
};
\addplot [red,only marks,thick,mark color=red,
mark=x] 
coordinates {
    (02, 71.848)
    (03, 71.93)
    (04, 72.08)
    (05, 72.28)
};
\addplot [red,only marks,thick,mark color=red,
mark=x] 
coordinates {
    (02, 71.71)
    (03, 72.11)
    (04, 72.31)
    (05, 72.21)
};
\addplot [red,only marks,thick,mark color=red,
mark=x] 
coordinates {
    (02, 71.86)
    (03, 72.12)
    (04, 72.12)
    (05, 72.23)
};

\addlegendentry{\hspace{-.0cm}\textbf{Degree}}
\addlegendentry{Degree 2 SP}
\addlegendentry{Degree 2 RP}
\addlegendentry{Degree 1 RP}

\end{axis}
\end{tikzpicture}}
	\caption{
           Training loss of models with varying numbers of layers across ten different seeds.
        }
	\label{fig:a1}
\end{figure}

\subsection{Ablation study - MNIST}
\label{section:ablation}
To assess the impact of our structured pruning strategy, we have also conducted an ablation study, on the MNIST task, considering both standard and increased function complexities. Specifically, we evaluated the performance of SP on the LogicNets baseline ($D=1$) and with a higher degree function ($D=2$). We evaluated SP performance on the LogicNets baseline ($D=1$) and with a higher degree function ($D=2$). Our results show that SP improves training loss and test accuracy over both the baseline and the more expressive functions, providing consistently higher accuracy results.

\begin{figure}[tbp]
	\centering
        \hspace{-0.5cm}
        \resizebox{0.9\columnwidth}{!}{\begin{tikzpicture}
\begin{axis}[width=1\columnwidth, height=0.9\columnwidth,ylabel={Test Accuracy (\%)},  grid=both, boxplot/draw direction=y, xtick={1,2,3,4}, xticklabel style={rotate=0},ylabel near ticks,
  xticklabels={D=1 RP, D=1 SP, D=2 RP, D=2 SP},]
    \addplot+[mark = *,fill,fill opacity=0.2,
    boxplot prepared={lower whisker=92.66, lower quartile=93.005, median=93.137, upper quartile=93.27, upper whisker=93.51},
    ] coordinates{};
    \addplot+[mark = *,fill,fill opacity=0.2,
    boxplot prepared={
      lower whisker=94.8, lower quartile=94.895, median=95.045, upper quartile=95.2125, upper whisker=95.33
    },
    ] coordinates{};
    \addplot+[mark = *,fill,fill opacity=0.2,
    boxplot prepared={lower whisker=95.51, lower quartile=95.545, median=95.621, upper quartile=95.705, upper whisker=95.94},
    ] coordinates{(0,95.27)};
    \addplot+[mark = *,fill,fill opacity=0.2,
    boxplot prepared={lower whisker=96.97, lower quartile=97, median=97.082, upper quartile=97.115, upper whisker=97.15}, 
    ] coordinates{(0,97.4)};

\addplot [red, smooth,<->]
coordinates {
    (2, 93.51)
    (2, 95.33)
}
node[
  text=red,
  font=\tiny,
  above,
  yshift=-0.9cm] at (0.5,94) [pos=0.7,font=\tiny]{$\mathbf{\Delta=1.82}$};
\addplot [brown, smooth,<->]
coordinates {
    (3, 93.51)
    (3, 95.94)
}
node[
  text=brown,
  font=\tiny,
  above,
  yshift=-0.9cm] at (0.5,94) [pos=0.7,font=\tiny]{$\mathbf{\Delta=2.43}$};
\addplot [black, smooth,<->]
coordinates {
    (4, 93.51)
    (4, 97.15)
}
node[
  text=black,
  font=\tiny,
  above,
  yshift=-0.9cm] at (0.5,94) [pos=0.7,font=\tiny]{$\mathbf{\Delta=3.64}$};
\end{axis}
\end{tikzpicture}}
	\caption{
            Test accuracy variation across ten seeds for D=1 and D=2 with random pruning (RP) and structured pruning (SP).
        }
	\label{fig:mnist}
\end{figure}
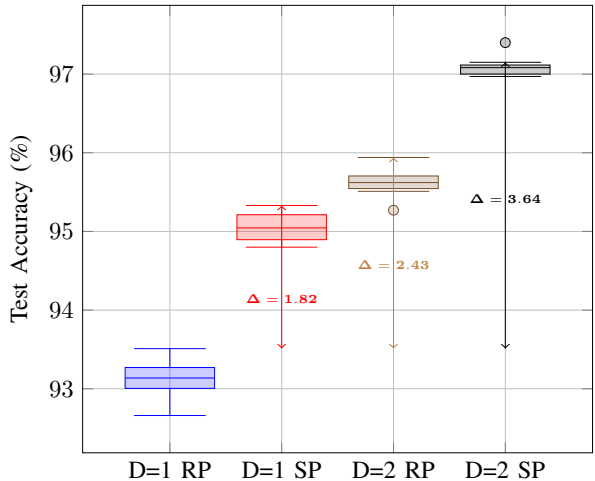

The leftmost box plot in Figure~\ref{fig:mnist} illustrates the performance of the baseline LogicNets (D=1) with random \textit{a priori} sparsity masks. A significant accuracy boost is observed in the next box plot, which replaces random pruning (RP) with our structured pruning (SP) algorithm. The rightmost two box plots demonstrate the same trend for the (D=2) networks. Thus, the pruning strategy is effective regardless of function expressivity, serving as a complementary solution that enhances accuracy only by learning the connectivity of the neural network.

The highlight of the experiment in Figure~\ref{fig:mnist} reveals that on the MNIST dataset, utilizing both higher-degree polynomial functions and our structured pruning algorithm can improve the accuracy by $3.64$ percentage points while significantly reducing the standard deviation between seeds.

\subsection{Comparison with prior work}
\label{section:evaluation}
We evaluate PolyLUT by considering the following metrics: accuracy, logic utilization, maximum frequency, and latency, with a strong focus on low latency for edge applications. Various model architectures were used to train PolyLUT, and to ensure a fair comparison, we selected parameters that give rise to comparable test accuracy with previous works such as LogicNets~\cite{logicnets}, FINN~\cite{finn}, \texttt{hls4ml}\cite{hls4ml}, Duarte \textit{et al.}\cite{duarte}, and Fahim \textit{et al.}\cite{fahim}, all while optimizing for latency and area utilization. NullaNet~\cite{nullanet} is not a direct comparison point as it employs floating-point for the first and last layers and reports hardware results only for the hidden layers. Moreover, LogicNets does not provide MNIST benchmarks.

\begin{table*}[htbp]
\caption{Evaluation of PolyLUT on multiple datasets. Bold indicates best in class.}
\begin{center}
\renewcommand{\arraystretch}{1.5} 
\begin{tabular}{clrrrrrrr}

&&\textbf{Accuracy}&\textbf{LUT}&\textbf{FF}&\textbf{DSP}&\textbf{BRAM}&\textbf{$\text{F}_\text{max}$ (MHz)}&\textbf{Latency (ns)}\\
\cellcolor[gray]{0.9}&\textbf{PolyLUT (NID Lite)}&\textbf{92.2\%}&\textbf{3165}&774&\textbf{0}&\textbf{0}&\textbf{580}&\cellcolor[gray]{0.9}\textbf{9}\\
\cline{2-9}
\cellcolor[gray]{0.9}&\textbf{LogicNets}\cite{logicnets}$^{\mathrm{a}}$&91.3\%&15949&1274&\textbf{0}&\textbf{0}&471&\cellcolor[gray]{0.9}13\\
\cline{2-9}
\multirow{-3}{*}{\cellcolor[gray]{.9}\textbf{UNSW-NB15}} &\textbf{Murovic \textit{et al.}}\cite{murovic}&\textbf{92.2\%}&17990&\textbf{0}&\textbf{0}&\textbf{0}&55&\cellcolor[gray]{0.9}18\\
\hline
\hline
\cellcolor[gray]{0.9}&\textbf{PolyLUT (HDR)}&\textbf{97.5\%}&\textbf{75131}&\textbf{4668}&\textbf{0}&\textbf{0}&\textbf{353}&\cellcolor[gray]{0.9}\textbf{17}\\
\cline{2-9}
\cellcolor[gray]{0.9}&\textbf{FINN}\cite{finn}&95.8\%&91131&-&\textbf{0}&5&200&\cellcolor[gray]{0.9}310\\
\cline{2-9}
\multirow{-3}{*}{\cellcolor[gray]{.9}\textbf{MNIST}} &\textbf{\texttt{hls4ml}}\cite{hls4ml}&95.0\%&260092&165513&\textbf{0}&\textbf{0}&200&\cellcolor[gray]{0.9}190\\
\hline\hline
\multirow{-1}{*}{\cellcolor[gray]{.9}\textbf{Jet substructure tagging}}&\textbf{PolyLUT (JSC-M Lite)}&\textbf{72.5\%}&\textbf{10169}&\textbf{631}&\textbf{0}&\textbf{0}&\textbf{598}&\cellcolor[gray]{0.9}\textbf{5}\\
\cline{2-9}
\multirow{-1}{*}{\cellcolor[gray]{.9}\textbf{(low accuracy)}} &\textbf{LogicNets}\cite{logicnets}$^{\mathrm{a}}$&71.8\%&37931&810&\textbf{0}&\textbf{0}&427&\cellcolor[gray]{0.9}13\\
\hline
\hline
\cellcolor[gray]{0.9}&\textbf{PolyLUT (JSC-XL)}&75.1\%&246071&12384&\textbf{0}&\textbf{0}&\textbf{203}&\cellcolor[gray]{0.9}\textbf{25}\\
\cline{2-9}
\multirow{-2}{*}{\cellcolor[gray]{.9}\textbf{Jet substructure tagging}}&\textbf{Duarte \textit{et al.}}\cite{duarte}&75.2\%&\multicolumn{2}{c}{88797$^{\mathrm{b}}$}&954&\textbf{0}&200&\cellcolor[gray]{0.9}75\\
\cline{2-9} 
\multirow{-2}{*}{\cellcolor[gray]{.9}\textbf{(high accuracy)}} &\textbf{Fahim \textit{et al.}}\cite{fahim}&\textbf{76.2\%}&\textbf{63251}&\textbf{4394}&38&\textbf{0}&200&\cellcolor[gray]{0.9}45\\

\multicolumn{9}{l}{$^{\mathrm{a}}$New results can be found on the LogicNets GitHub page.}\\
\multicolumn{9}{l}{$^{\mathrm{b}}$Paper reports ``LUT+FF".}\\
\end{tabular}
\renewcommand{\arraystretch}{1.5}
\label{table:evaluation}
\end{center}

\end{table*}

\subsubsection{Network intrusion detection}
We evaluate PolyLUT for the network intrusion detection task, comparing our results with LogicNets~\cite{logicnets} and Murovic \textit{et al.}\cite{murovic}. The outcomes are summarized in Table~\ref{table:evaluation}.

For this task, we utilized the NID Lite architecture from Table~\ref{table:networks} with a polynomial degree set to $4$. PolyLUT achieves higher accuracy than LogicNets and matches the accuracy by Murovic \textit{et al.} In terms of latency, PolyLUT demonstrates a $1.44\times$ reduction compared to LogicNets and a $2\times$ compared to Murovic \textit{et al.}. Furthermore, PolyLUT significantly reduces the LUT count, with a $5.04\times$ creduction compared to LogicNets and a $5.68\times$ reduction compared to Murovic \textit{et al.}.

\subsubsection{Handwritten digit recognition}
To assess PolyLUT's performance on the MNIST digit classification dataset, we compare it with FINN's SFC-max model~\cite{finn}, a fully unfolded implementation, and the ternary neural network (TNN) used in \texttt{hls4ml}\cite{hls4ml}. The evaluation results are detailed in Table~\ref{table:evaluation}. 

For this task, we employed the HDR architecture from Table~\ref{table:networks} and trained the network with a polynomial degree of $4$. PolyLUT achieves significantly higher accuracy compared to both \texttt{hls4ml} and FINN. PolyLUT demonstrates a substantial latency reduction of $18.24\times$ against FINN and a latency reduction of $11.18\times$ against \texttt{hls4ml}. In terms of resource utilization, PolyLUT completely eliminates block random-access memory (BRAM) utilization compared to FINN and reduces LUT utilization by $1.21\times$. Against \texttt{hls4ml}, PolyLUT achieves a $3.46\times$ reduction in LUT utilization.

These significant latency reductions are achieved without compromising accuracy, through our effective handling of function complexity and the minimization of exposed datapaths, which are prone to causing bottlenecks.

\subsubsection{Jet substructure tagging}
We evaluated our method on the jet substructure tagging task by comparing it against three existing approaches: LogicNets~\cite{logicnets}, Duarte \textit{et al.}~\cite{duarte}, and Fahim \textit{et al.}~\cite{fahim}. Our comparison is divided into two sections, with one section of the table focusing on lower latency as per LogicNets. Our method demonstrates superior performance in terms of latency, surpassing all previous approaches, thus highlighting the effectiveness of our methodology for ultra-fast processing applications.

For our experiments, we utilized two architectures: JSC-M Lite and JSC-XL, with polynomial degrees set to $6$ and $4$, respectively, as outlined in Table~\ref{table:networks}. Table~\ref{table:evaluation} provides a comprehensive comparison of our work. 

Our JSC-M Lite surpasses LogicNets in terms of accuracy while also reducing the LUT utilization by $3.73\times$ and the latency by $2.6\times$. Our JSC-XL architecture reaches comparable accuracy with Duarte \textit{et al.} while reducing the latency by a factor of $3\times$. In comparison to Fahim \textit{et al.}, due to their higher precision (fixed-point), our method exhibits a decrease in accuracy of $1.1$ percentage point, however, we achieve a significant improvement in latency by a factor of $1.8\times$, which may be critical to particle acceleration.

Unlike Duarte \textit{et al.} and Fahim \textit{et al.}, our approach does not use DSP blocks, though this can result in a higher LUT count in specific cases where both high neuron fan-in and high precision are required to achieve a given accuracy. This is exemplified in the JSC-XL architecture.

\section{Conclusion and Further Work}
In this work, we introduced PolyLUT, a novel DNN-hardware co-design methodology tailored to meet the stringent demands of on-edge applications, particularly in terms of ultra-low latency and minimal area requirements. Our approach involves mapping sparse and quantized polynomial neural networks to netlists of LUTs by training multivariate polynomials instead of linear functions. This method leverages the inherent capabilities of LUTs to represent any function, allowing us to maintain accuracy with shallower neural network models.

By reducing stochasticity and improving performance consistency, the hardware-aware structured pruning enhances the reliability of neural network evaluations and optimizations, making it a more robust choice for neural architecture search and other applications where precision and stability are critical.

We have demonstrated the effectiveness of our approach on three different datasets: network intrusion detection, handwritten digit classification, and jet substructure tagging. Our method achieved significant latency improvements over prior works, with reductions of $2\times$, $11.18\times$, and $2.6\times$ for these tasks, respectively, while maintaining similar accuracies.

Despite the polynomial scaling of training parameters for fixed $D$, the L-LUT size scales exponentially, which limits the network architectures to having few inputs of low precision. Therefore, future work could focus on tackling the scaling up these lookup tables from both software or hardware perspectives. From a software perspective, this could be done by redesigning the structure of the neural network, and from a hardware perspective, through the utilization of \textit{don't cares}.

Another clear extension of our work would be to incorporate neural architecture search (NAS), enabling the exploration of optimal architectures and hyperparameters such as bit-width, fan-in, and degree. The parameters in Table~\ref{table:networks} were determined through manual tuning. NAS could potentially identify architectures that offer even better performance and efficiency, further enhancing the capabilities of our methodology.

\bibliographystyle{IEEEtran}
\begingroup
\raggedright
\bibliography{bibs}
\endgroup

\end{document}